# A Robust Multilabel Method Integrating Rule-based Transparent Model, Soft Label Correlation Learning and Label Noise Resistance

Qiongdan Lou, Zhaohong Deng, *Senior Member, IEEE*, Qingbing Sang, Zhiyong Xiao, Kup-Sze Choi, *Senior Member, IEEE*, Shitong Wang

*Abstract*—Model transparency, label correlation learning and robustness to label noise are crucial for multilabel learning. However, few existing methods consider these three characteristics simultaneously. To address this challenge, we propose the robust multilabel Takagi-Sugeno-Kang fuzzy system (R-MLTSK-FS) with three mechanisms. First, we design a soft label learning mechanism to reduce the effect of label noise by explicitly measuring the interactions between labels, which is also the basis of the other two mechanisms. Second, rule-based multi-output TSK FS is used as the base model to efficiently model the inference relationship between features and soft labels in a more transparent way than many existing multilabel models. Third, to further improve the performance of multilabel learning, we build a correlation enhancement learning mechanism based on the soft label space and the fuzzy feature space. Extensive experiments are conducted to demonstrate the superiority of the proposed method.

## I. Introduction

MULTILABEL learning concerns instances that can be associated with more than one labels. For example, an article can be labeled as being related to "politics", "culture" and "religion" at the same time; and a travel photo can be given the labels "beach", "sunrise", "sail" and "tourist" simultaneously because of the presence of the corresponding objects. For multilabel learning, label correlation learning, model transparency and robustness against label noise are essential for the following reasons. Constructing the correlation between labels is the basic work to improve the performance of multilabel learning [1, 2]; transparent structure is important to enhance the interpretability of multilabel learning [3]; and robustness against label noise can enhance the effectiveness in practical applications under noisy environment [4]. However, these three desirable characteristics are rarely studied simultaneously in multilabel learning, which deserve in-depth investiations.

Based on the above analysis, we aim to develop a multilabel learning method with label correlation learning ability and strong fuzzy inference ability even under the influence of noisy labels. To achieve the goal, a multilabel learning classifier called robust multilabel Takagi-Sugeno-Kang fuzzy system, i.e., R-MLTSK-FS, is proposed by developing three enabling mechanisms. The first mechanism concerns *soft label learning*. The R-MLTSK-FS maps the original label matrix to the soft label space where each soft label is affected by all the original labels. The mechanism thus reduces the influence of label noise in the original label space, and forms the basis of the other two mechanisms. The second mechanism concerns the *construction of soft multilabel loss function*. In R-MLTSK-FS, the IF-THEN rule-based multi-output Takagi-Sugeno-Kang fuzzy system (TSK FS) is used to model the inference between the inputs and outputs. Specifically, the IF-part of a multi-output TSK FS is leveraged to transform the original feature matrix into the fuzzy feature space; the THEN-part is used to implement the inference between inputs and outputs; and the regression loss is constructed based on the multi-output TSK FS and soft label learning. The adoption of multi-output TSK FS is advantageous in that the rule-based multi-output TSK FS makes the proposed R-MLTSK-FS more transparent than traditional models. In addition, the nonlinear learning ability of multi-output TSK FS can also be utilized to learn the mapping relationship between inputs and outputs. The third mechanism concerns *correlation enhancement learning*. The mechanism establishes associations between any two soft labels and their corresponding fuzzy discriminative features, which can effectively improve the performance of R-MLTSK-FS.

In essence, the multi-output TSK FS adopted in this paper is a kind of interpretable fuzzy model [92, 93]. Because of the transparent and efficient rule inference ability, we combine the multi-output TSK FS with multilabel learning. That is, based on the transparent rule inference structure and the constraint terms of multilabel learning, a multilabel method integrating rule-based transparent model, soft label correlation learning and label noise resistance is proposed in this paper. On the whole, the main contributions of this paper are summarized as follows:

(1) A robust multilabel TKS FS with transparency, label correlation learning ability and robustness against label noise, is proposed in this paper.

This work was supported in part by the National key R & D plan (2022YFE0112400), the Chinese Association for Artificial Intelligence (CAAI)-Huawei MindSpore Open Fund under Grant CAAIXSJLJJ-2021-011A, the NSFC (62176105), the Hong Kong Research Grants Council (PolyU 152006/19E), the Project of Strategic Importance of the Hong Kong Polytechnic University (1-ZE1V). (Corresponding author: Zhaohong Deng).

Q. Lou, Z. Deng, Q. Sang, Z. Xiao, S. Wang are with the School of Artificial Intelligence and Computer Science, Wuxi 214122, China (e-mail: 6171610005@stu.jiangnan.edu.cn; dengzhaohong@jiangnan.edu.cn; sangqb@163.com; zhiyong.xiao@jiangnan.edu.cn; wxwangst@aliyun.com).

K.S. Choi is with the Centre for Smart Health, Hong Kong Polytechnic University. (e-mail: kschoi@ieee.org).

(2) A soft label learning mechanism is constructed to explicitly measure the interaction between labels and reduce the influence of label noise.

(3) A soft multilabel loss function is constructed based on soft labels and multi-output TSK FS to model the nonlinear mapping relationship between features and soft labels to improve the learning performance of R-MLTSK-FS.

(4) A correlation enhancement learning mechanism based on soft label space and fuzzy feature space is built to strengthen the associations between any two soft labels and their corresponding fuzzy discriminative features, thus further enhancing the learning ability of R-MLTSK-FS.

(5) The IF-THEN rule-based multi-output TSK FS is utilized to model the inference between inputs and outputs, so as to improve the transparency of the learning process of R-MLTSK-FS.

(6) Extensive experiments are conducted using 10 benchmark multilabel datasets and 3 synthetic multilabel datasets to compare with 11 existing methods. Comprehensive evaluations are carried out by classification performance assessment, robustness analysis, effectiveness analysis of soft label learning and correlation enhancement learning, parameter analysis, convergence analysis, statistical analysis, running time analysis and interpretability analysis.

The rest of this paper is organized as follows. Section II reviews the concepts of multilabel learning, related work of multilabel learning, and the traditional TSK FS. Section III gives details of the proposed method. Extensive experimental analyses are presented and discussed in Section IV. Finally, Section V summarizes the paper.

## II. BACKGROUND KNOWLEDGE

In this section, the problem statement and related work of the multilabel learning research concerned in the study is given, followed by a review of traditional TSK FS.

### A. Problem Statement

Let $\mathcal{X} \in \mathcal{R}^D$ and $\mathcal{Y} \in \mathcal{R}^L$ be a $D$-dimensional feature space and an $L$-dimensional label space respectively. $\mathcal{D} = \{(x_i, y_i)\}_{i=1}^N$ is the training set with $N$ samples. $X = [x_1, x_2, ..., x_N] \in \mathcal{R}^{D \times N}$ is the input matrix, and $Y = [y_1, y_2, ..., y_N] \in \mathcal{R}^{L \times N}$ is the output matrix. In multilabel learning, the label of an instance $x_i = [x_{i1}, x_{i2}, ..., x_{iD}]^T$ is given by a vector $y_i = [y_{i1}, y_{i2}, ..., y_{iL}]^T$. If $x_i$ is related to the $j$th label, then $y_{ij} = 1$, otherwise, $y_{ij} = 0$. The aim of this study is to find a robust mapping function $f: \mathcal{X} \to \mathcal{Y}$ that can reduce the influence of label noise and effectively predict the label vector for a new instance on the basis of transparent inference rules.

### B. Related Work of Multilabel Learning

For label correlation learning, existing multilabel methods are mainly based on first-order [5], second-order [6] and high-order [7] strategies to consider the correlation between labels. *First-order methods* ignore label correlation and adopt label-by-label approach for multilabel learning. For example, sparse weighted instance-based multilabel (SWIM) realizes multilabel learning only based on the association between instances [8]. Domain knowledge is utilized in [72] to improve the performance of multilabel classifiers. Meta multi-instance multilabel learning (MetaMIML) constructs multiple label learners in a parallel way [73]. *Second-order methods* build pairwise relationship between labels. For example, labels related to a sample are ranked before labels unrelated to the sample [9]. Multilabel learning with global and local label correlation (GLOCAL) decomposes the Laplacian matrix to indirectly learn the correlation between any two labels [10]. Inconsistent rankers are analyzed in both instance-oriented preference distribution learning (IPDL) and ranker-oriented preference distribution learning (RPDL) to learn the ranking between any two labels [74]. Distributed information-theoretic semisupervised multilabel learning (dITS$^2$ML$^2$) builds the correlation between any two labels through some anchor data [75]. The co-occurrence ratio between two labels is calculated in [76] to model the label correlation. *High-order methods* construct the correlation between multiple labels simultaneously. For example, cross-coupling aggregation (COCOA) first models the correlation between random label pairs and then aggregates their learning effects [11]. Multilabel classification with label-specific features and label-specific classifiers (MLC-LFLC) introduces sparse learning to analyze the dependency between a single label and other labels [12]. Multilabel classification with group-based mapping (MC-GM) first divides the samples into multiple groups and learns the instance-group correlation and group-label correlation for each group, and then establishes the inter-group correlation to facilitate the learning process of the predictor [77]. To establish the label correlation learning in graph convolutional networks (GCNs), correlation matrix based on the co-occurrence ratio between labels is constructed in both classifier learning GCN (C-GCN) and prediction learning GCN (P-GCN) to guide the information propagation [78]. To analyze the correlation between labels, non-aligned incomplete multiview and missing multilabel learning method (NAIM$^3$L) characterizes the local and global structures of multiple labels as low-rank and high-rank respectively [79]. High-order correlation between labels is analyzed in [80] through joint label-specific classifiers training.

For model transparency in multilabel learning, existing work is mainly based on rules or logical inference to achieve transparency [13]. For example, as a hierarchical multilabel learning method, hierarchical multilabel classification with a genetic algorithm (HMC-GA) [14] utilizes genetic algorithm to induce classification rules for protein function prediction. The gradient-weighted class activation mapping (Grad-CAM) is used in [15] to realize inferential interpretation for predicted label results. Causal discovery is exploited in [16] to analyze the specific features of a label. A transparent multilabel elastic-net classifier is utilized in [81] to select essential features for discriminating labels. Interpretable deep learning (IDeL) attempts to demonstrate transparency by designing feature penetration, instance aggregation and feature perturbation for black-box models [82]. The multilabel Takagi-Sugeno-Kang fuzzy system, i.e., ML-TSK FS [17] offers good transparency through fuzzy rule-based structure and fuzzy inference. Among the above existing multilabel methods, ML-TSK FS has shown more promising performance because it realizes a complete inference process from feature to label.

For robustness against label noise, considerable studies have been conducted because of the need to realize practical application [18, 19]. For example, class-conditional multilabel noise (CCMN) [20] designs two unbiased estimators with error bounds to reduce the influence of label noise. Multilabel noise robust collaborative learning (RCML) [21] employs group lasso to detect noisy labels. Partial multilabel learning with noisy label identification (PML-NI) [22] builds the feature-induce noise term to identify noisy labels. Multilabel iterated learning (MILe) [23] tackles learning bottleneck in successive generations of teacher and student networks to improve the robustness against label noise. Noisy label tolerated partial multilabel learning (NATAL) [24], instead of removing noisy labels directly, reduces the impact of noisy labels by assuming that the label information is precise and the feature information is inadequate. A self-representation model with a low-rank constraint is employed in [83] to reduce the influence of label noise. While preserving the structure and relationships in the original label space, multilabel feature instance label selection (mFILS) decomposes the label space into a low-dimensional space to reduce the interference of label noise [84]. By adopting different sampling priors and loss functions, [85] constructs a heterogeneous co-learning framework to rectify noisy labels. A label corruption matrix estimated on trusted data is designed in [86] to correct the loss, so as to enhance the robustness against label noise. Self-adaptive training, early-learning regularization and joint co-regularized training are investigated in [87] to improve the robustness against label noise. Global-local label correlation (GLC) decomposes the label matrix into a noise label matrix and a ground-truth label matrix, and then exploits the global label joint correlations to eliminate the impact of label noise [88]. Partial multilabel learning based on sparse asymmetric label correlations (PML-SALC) constructs the sparse asymmetric label correlation matrix to reduce the influence of label noise [89].

The research studies discussed above indicates that the importance of label correlation, robustness against noisy labels and model transparency in multilabel learning. However, these three characteristics are rarely taken into account at the same time and it is therefore necessary to carry in-depth research accordingly.

*C. TSK Fuzzy System*

While inference ability can be improved by various approaches [69, 70, 71], TSK FS is a classical and recognized inference model based on fuzzy rules with superior interpretability (transparency) and learning ability. It has been successfully applied in different areas, e.g., transfer learning [25, 26], multiview learning [27], multitask learning [28] and others [29, 30, 31, 32]. For a TSK FS with $K$ rules, the $k$th rule can be expressed as follows:

IF: $x_1$ is $A_1^k \wedge x_2$ is $A_2^k \wedge ... \wedge x_D$ is $A_D^k$,

THEN: $f^k(\boldsymbol{x}) = c_0^k + c_1^k x_1 + \cdots + c_D^k x_D$,
$k = 1, 2, ..., K$ (1)

where $D$ is the feature dimension, and $f^k(\boldsymbol{x})$ is the output of instance $\boldsymbol{x}$ on the $k$th rule. $A_d^k$ ($d = 1, 2, ..., D$) in the IF-part is the antecedent fuzzy set, which can be described by membership functions. $c_d^k$ in the THEN-part is the consequent parameter.

Depending on application scenarios, different membership functions can be chosen for the antecedent fuzzy sets. Gaussian function, which is commonly used, is adopted in this paper and the corresponding membership function associated with $A_d^k$ can be expressed as follows:

$$\mu_{A_d^k}(x_d) = \exp\left\{-\frac{1}{2}(\frac{x_d - m_d^k}{\delta_d^k})^2\right\} \quad (2)$$

where $m_d^k$ and $\delta_d^k$ can be obtained using different methods. In the absence of domain knowledge, data-driven methods are usually utilized to estimate $m_d^k$ and $\delta_d^k$. For example, the Var-Part clustering has been used for this purpose [33]. It is insensitive to the parameters and is therefore beneficial in terms of stability and practicability. Hence, the Var-Part clustering is used in this study.

For TSK FS, the firing strength of instance $\boldsymbol{x}$ on the $k$th rule can be computed as follows:

$$\mu^k(\boldsymbol{x}) = \prod_{d=1}^{D} \mu_{A_d^k}(x_d) \quad (3)$$
$$\tilde{\mu}^k(\boldsymbol{x}) = \mu^k(\boldsymbol{x})/\sum_{k'=1}^{K} \mu^{k'}(\boldsymbol{x}) \quad (4)$$

where Eq. (4) is the normalized form of Eq. (3).

Finally, the output of TSK FS for instance $\boldsymbol{x}$ can be expressed as

$$y = f(\boldsymbol{x}) = \sum_{k=1}^{K} \tilde{\mu}^k(\boldsymbol{x}) f^k(\boldsymbol{x}) \quad (5)$$

In fact, Eq. (5) can also be expressed as a linear model in the new fuzzy feature space, that is,

$$y = f(\boldsymbol{x}) = \boldsymbol{c}^T \boldsymbol{x}_g \quad (6)$$

where

$$\boldsymbol{x}_e = [1, \boldsymbol{x}^T]^T \in \mathcal{R}^{(D+1) \times 1} \quad (7)$$
$$\tilde{\boldsymbol{x}}^k = \tilde{\mu}^k(\boldsymbol{x}) \boldsymbol{x}_e \in \mathcal{R}^{(D+1) \times 1} \quad (8)$$
$$\boldsymbol{x}_g = [(\tilde{\boldsymbol{x}}^1)^T, (\tilde{\boldsymbol{x}}^2)^T, ..., (\tilde{\boldsymbol{x}}^K)^T]^T \in \mathcal{R}^{K(D+1) \times 1} \quad (9)$$
$$\boldsymbol{c}^k = [c_0^k, c_1^k, ..., c_D^k]^T \in \mathcal{R}^{(D+1) \times 1} \quad (10)$$
$$\boldsymbol{c} = [(\boldsymbol{c}^1)^T, (\boldsymbol{c}^2)^T, ..., (\boldsymbol{c}^K)^T]^T \in \mathcal{R}^{K(D+1) \times 1} \quad (11)$$

Here, $\boldsymbol{x}_g$ is the fuzzy representation of instance $\boldsymbol{x}$ in the new feature space generated by fuzzy rules. $\boldsymbol{c}$ is the consequent parameter vector of all the rules, which can be optimized by solving the linear model in Eq. (6).

III. PROPOSED METHOD: R-MLTSK-FS

*A. System Architecture*

The architecture of R-MLTSK-FS proposed in this study is shown in Fig. 1. It aims to develop a robust multilabel model with fuzzy inference ability, label correlation learning ability and resistance against noisy labels. R-MLTSK-FS contains three mechanisms for soft label learning, soft multilabel loss function construction and correlation enhancement learning, respectively.

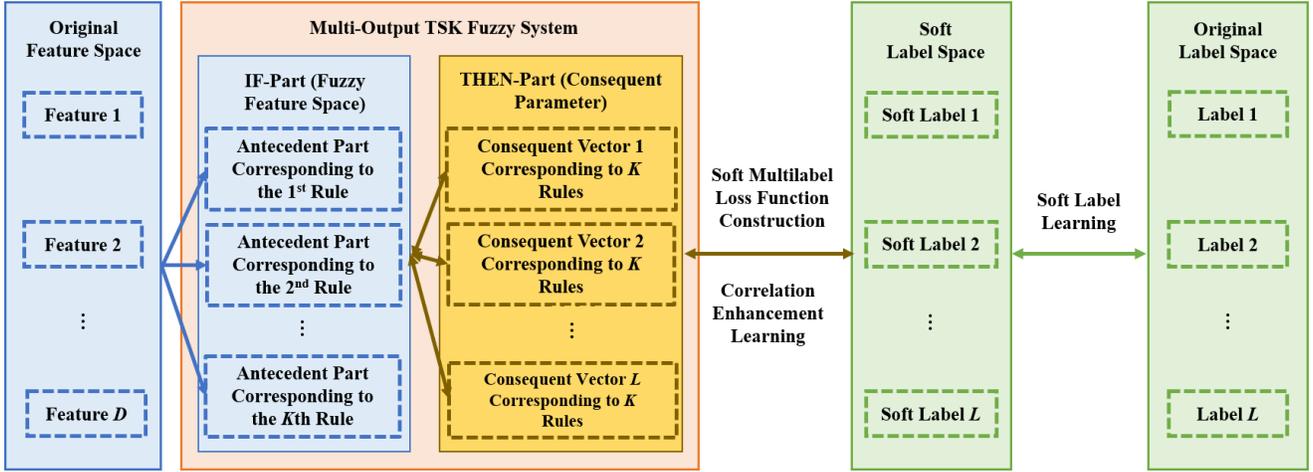

Fig. 1 The architecture of the proposed R-MLTSK-FS.

The first mechanism, soft label learning, maps the original labels to a soft label space by linear transformation. Each soft label in the soft label space is associated with all the original labels, which reduces the influence of label noise in the original label space. It forms the basis of the other two mechanisms. The second mechanism, i.e., soft multilabel loss function construction, leverages the IF-part of the multi-output TSK FS to transform the original features into the fuzzy feature space, uses the THEN-part of the multi-output TSK FS to complete the inference between inputs and outputs, and then constructs the regression function between the fuzzy feature space and the soft label space. The rule-based multi-output TSK FS makes R-MLTSK-FS transparent in modeling inference relationship between features and labels. The third mechanism, correlation enhancement learning, implements label correlation learning by establishing the associations between any two soft labels and their corresponding fuzzy discriminative features. This mechanism further enhances the learning ability of R-MLTSK-FS.

The details of R-MLTSK-FS are expanded in the following three sections. The learning criteria of R-MLTSK-FS is introduced in Section III-B. The optimization process is given in Section III-C. The algorithm is presented in Section III-D, and the computational complexity is analyzed in Section III-E.

### B. Learning Criteria of R-MLTSK-FS

According to the analysis in Section III-A, the multilabel learning problem in this paper can be expressed as the following optimization objective criteria:

$$\min_{S,C} \beta \cdot Sof\_lab(S) + Sof\_los(S,C) + \gamma \cdot Cor\_enh(S,C) \quad (12)$$

The first term represents soft label learning, where $S$ transforms the original labels to the soft labels. The second term represents soft multilabel loss function construction, where $C$ is used to predict the labels from the fuzzy feature space to the soft label space. The third term represents correlation enhancement learning, which is used to measure the association between any two soft labels and their corresponding fuzzy discriminative features. The hyperparameters $\beta$ and $\gamma$ are used to balance the influences of different terms in Eq. (12). The solutions of $S$ and $C$ can be obtained by optimizing Eq. (12). The implementation of the three terms is described below.

*1) Soft Label Learning based on Original Label Space and Soft Label Space*

To learn the $l$th label effectively, the interference of label noise in $Y_l \in \mathcal{R}^{1 \times N}$ ($1 \leq l \leq L$) (i.e., the $l$th row in $Y$) can be reduced by considering the influence of all the labels on $Y_l$ comprehensively. Based on this, we develop the soft label learning mechanism and assume that each label is associated with all the other original labels to some extent. The learning process involves two steps. First, we construct the label transformation $\phi_1$ to effectively measure the interaction between the labels. $\phi_1$ maps the output matrix $Y$ explicitly from the original label space to the soft label space. In the soft label space, each soft label is associated with all the original labels. The transformation function of $\phi_1$ is defined as:

$$\phi_1(S) = SY \quad (13)$$

where $S = [s_1, s_2, \ldots, s_L]^T \in \mathcal{R}^{L \times L}$, and $s_l \in \mathcal{R}^{L \times 1}$ ($1 \leq l \leq L$) represents the $l$th soft label learning vector. For the $l$th label $Y_l$ in the original label space, the corresponding soft label vector in the soft label space is $s_l^T Y \in \mathcal{R}^{1 \times N}$ ($1 \leq l \leq L$).

Second, the loss function of soft label learning is constructed by minimizing the difference between the original label matrix and the soft label matrix to ensure the classification performance. Therefore, the overall soft label learning is defined as:

$$\min_{S} Sof\_lab(S)$$
$$= \min_{S} \sum_{i=1}^{N} \left\| (Y_{,i} - SY_{,i})^T \right\|_{2,1}$$
$$= \min_{S} \left\| (Y - SY)^T \right\|_{2,1} \quad (14)$$

where $Y_{,i} \in \mathcal{R}^{L \times 1}$ represents the $i$th column of $Y$, that is, the label vector of the $i$th training instance.

Although different regularization norms can be used in Eq. (14), we choose the $L_{2,1}$ norm for two reasons: (1) since $L_{2,1}$ norm has the characteristic of row sparsity, we can screen out the original label subsets which have significant impact on the corresponding soft label, (2) L$_{2,1}$ norm is well-known for its ability in robust group selection [34, 35, 36], which is helpful to reduce the impact of label noise on soft label learning.

*2) Soft Multilabel Loss Function Construction based on TSK FS*

Multilabel loss function can be constructed by employing an evaluation metric as the multilabel objective function [37, 38], or by using linear regression to derive the multilabel loss function [39, 40, 41]. Unlike these methods, we construct the loss function using soft label learning and multi-output TSK FS, which essentially constructs a rule-based transparent model that maps the original feature space to the soft label space. The construction of the soft multilabel loss function is divided into three steps. First, the original feature matrix is transformed into the fuzzy feature space using the IF-part of the fuzzy rules. Second, the inference between inputs and outputs is obtained using the THEN-part of fuzzy rules. Third, the regression loss function is constructed based on the fuzzy rules and soft labels.

To clarify the difference between the multi-output TSK FS used in this paper and the traditional single-output TSK FS in Eq. (1), we first give the structure of the multi-output TSK FS. Specifically, for the multi-output TSK FS with $K$ rules, the $k$th rule is expressed as follows:

IF: $x_1$ is $A_1^k \wedge x_2$ is $A_2^k \wedge ... \wedge x_D$ is $A_D^k$,
THEN: $f_1^k(\boldsymbol{x}) = c_{10}^k + c_{11}^k x_1 + \cdots + c_{1D}^k x_D$,
$$\vdots$$
$$f_l^k(\boldsymbol{x}) = c_{l0}^k + c_{l1}^k x_1 + \cdots + c_{lD}^k x_D,$$
$$\vdots$$
$$f_L^k(\boldsymbol{x}) = c_{L0}^k + c_{L1}^k x_1 + \cdots + c_{LD}^k x_D,$$
$$1 \leq k \leq K \text{ and } 1 \leq l \leq L \quad (15)$$

where $D$ is the feature dimension. $A_d^k$ ($d = 1, 2, ..., D$) represents the antecedent fuzzy set, which is the same as the antecedent fuzzy set of the single-output TSK FS in Eq. (1). $f_l^k(\boldsymbol{x})$ is the $l$th output of the multi-output TSK FS under the $k$th rule for the instance $\boldsymbol{x}$. $\boldsymbol{c}_l = [(\boldsymbol{c}_l^1)^T, (\boldsymbol{c}_l^2)^T, ..., (\boldsymbol{c}_l^K)^T]^T \in \mathcal{R}^{K(D+1)\times 1}$ is the $l$th consequent parameter vector corresponding to the $l$th output, and $\boldsymbol{c}_l^k = [c_{l0}^k, c_{l1}^k, ..., c_{lD}^k]^T \in \mathcal{R}^{(D+1)\times 1}$. Therefore, the consequent parameter matrix composed of $L$ consequent parameter vectors is $\boldsymbol{C} = [\boldsymbol{c}_1, \boldsymbol{c}_2, ..., \boldsymbol{c}_L]^T \in \mathcal{R}^{L\times K(D+1)}$. The soft multilabel loss function can be constructed. The details are as follows.

- **IF-part implementation of fuzzy rules.** In the multi-output TSK FS with $K$ rules, the fuzzy feature matrix obtained by $\boldsymbol{X}$ using fuzzy rules is given by
$$\boldsymbol{X}_g = [\boldsymbol{x}_{g,1}, \boldsymbol{x}_{g,2}, ..., \boldsymbol{x}_{g,N}] \in \mathcal{R}^{K(D+1)\times N} \quad (16)$$
where $\boldsymbol{x}_{g,i}$ ($1 \leq i \leq N$) is mapped by the instance $\boldsymbol{x}_i$ through the IF-part of fuzzy rules, and it can be obtained by Eqs. (2)-(4) and (7)-(9).

  Compared with the original features, the rule-based fuzzy features can empower R-MLTSK-FS to analyze the implicit inference relationship between features and labels [42], thereby strengthening the learning ability.

- **THEN-part adaptation of fuzzy rules.** Based on Eq. (6), the THEN-part of multi-output TSK FS is used to complete the inference. Therefore, we construct the inference function $\phi_2$ to complete the rule structure. That is,
$$\phi_2(\boldsymbol{C}) = \boldsymbol{C}\boldsymbol{X}_g \quad (17)$$
Corresponding to the $l$th output in multi-output TSK FS, the consequent parameter vector $\boldsymbol{c}_l$ ($1 \leq l \leq L$) in $\boldsymbol{C}$ is used to learn the $l$th soft label.

- **Construction of regression loss.** The loss function is a fundamental part of the optimization objective for multilabel classification. In this paper, it is built based on soft label learning and multi-output TSK FS. Combining Eqs. (13) and (17), we construct the soft multilabel loss function as follows:
$$\min_{\boldsymbol{S},\boldsymbol{C}} Sof\_los(\boldsymbol{S}, \boldsymbol{C})$$
$$= \min_{\boldsymbol{S},\boldsymbol{C}} \|(\boldsymbol{S}\boldsymbol{Y} - \boldsymbol{C}\boldsymbol{X}_g)^T\|_{2,1} + \alpha\|\boldsymbol{C}\|_F^2 \quad (18)$$
where $\alpha$ is a hyperparameter to balance the influence of the soft multilabel loss function and the regularization term. Taking the Frobenius norm $\|\cdot\|_F$ as the regularization term not only reduces the risk of overfitting, but also facilitates the solution of consequent parameter matrix $\boldsymbol{C}$.

*3) Correlation Enhancement Learning based on Soft Label Space and Fuzzy Feature Space*

Section I has clarified that mining the correlation information between labels can effectively improve the performance of the model. Based on this, we utilize the correlation information on the basis of soft label learning and fuzzy features as follows:

$$\min_{\boldsymbol{S},\boldsymbol{C}} Cor\_enh(\boldsymbol{S}, \boldsymbol{C})$$
$$= \min_{\boldsymbol{S},\boldsymbol{C}} \sum_{i=1}^{L}\sum_{j=1}^{L} \|(\boldsymbol{s}_i^T\boldsymbol{Y} - \boldsymbol{s}_j^T\boldsymbol{Y})^T\|^2 \boldsymbol{c}_i^T\boldsymbol{c}_j \quad (19)$$

where $\boldsymbol{s}_l^T\boldsymbol{Y} \in \mathcal{R}^{1\times N}$ ($1 \leq l \leq L$) represents the $l$th soft label vector corresponding to $N$ samples. $\boldsymbol{c}_l \in \mathcal{R}^{K(D+1)\times 1}$ ($1 \leq l \leq L$) is used to learn the discriminative features from fuzzy feature space for the $l$th soft label.

The objective function in Eq. (19) is developed based on three arguments: 1) $\boldsymbol{c}_l$ ($1 \leq l \leq L$) can be used to learn the discriminative features from the fuzzy feature space of the $l$th soft label. 2) The information expressed by soft labels is basically consistent with that of original labels. 3) The correlation between two labels is consistent with the correlation between their discriminative features. For example, there is an intersection between the labels "Cat" and "Animal", and then their discriminative features should partially overlap. Similarly, the larger the difference between two labels, the smaller the similarity between their discriminative features.

Hence, refer to Eq. (19), the larger the difference between the $i$th and $j$th soft labels, the more significant the difference between their fuzzy discriminative features, and further, the smaller the value of $\boldsymbol{c}_i^T\boldsymbol{c}_j$. Further, Eq. (19) can be expressed as:

$$\min_{\boldsymbol{S},\boldsymbol{C}} Cor\_enh(\boldsymbol{S}, \boldsymbol{C})$$

$$= \min_{S,C} \sum_{i=1}^{L}\sum_{j=1}^{L}\|(s_i^T Y - s_j^T Y)^T\|^2 c_i^T c_j$$
$$= \min_{S,C} 2\text{Tr}(Y^T S^T L S Y) \tag{20}$$

where $L = D - R$, $R = CC^T \in \mathcal{R}^{L \times L}$, $D \in \mathcal{R}^{L \times L}$ is a diagonal matrix, and $D_{ii} = \sum_{j=1}^{L} R_{ij}$.

### C. Complete Objective Function and its Optimization

By integrating Eqs. (14), (18) and (20), the multilabel learning problem in Eq. (12) is defined, and the complete objective function of R-MLTSK-FS is expressed as:

$$\min_{S,C} \beta \cdot Sof\_lab(S) + Sof\_los(S,C) +$$
$$\gamma \cdot Cor\_enh(S,C)$$
$$= \min_{S,C} \beta\|(Y - SY)^T\|_{2,1} + \|(SY - CX_g)^T\|_{2,1} +$$
$$\alpha\|C\|_F^2 + 2\gamma\text{Tr}(Y^T S^T L S Y)$$
$$= \min_{S,C} \|(SY - CX_g)^T\|_{2,1} + \alpha\|C\|_F^2 + \beta\|(Y - SY)^T\|_{2,1} +$$
$$2\gamma\text{Tr}(Y^T S^T L S Y) \tag{21}$$

To optimize $S$ and $C$, we adopt an alternating direction minimization strategy which divides Eq. (21) into two subproblems, namely, the $S$-subproblem and the $C$-subproblem. The optimization processes are as follows.

#### 1) $S$-Subproblem

By fixing $C$, the $S$-subproblem can be expressed as:

$$S^* = argmin_S \|(SY - CX_g)^T\|_{2,1} + \beta\|(Y - SY)^T\|_{2,1} +$$
$$2\gamma\text{Tr}(Y^T S^T L S Y) \tag{22}$$

In Eq. (22), the Lagrange function for $S$ is

$$L(S) = \|(SY - CX_g)^T\|_{2,1} + \beta\|(Y - SY)^T\|_{2,1} +$$
$$2\gamma\text{Tr}(Y^T S^T L S Y) \tag{23}$$

Set the derivative of Eq. (23) with respect to $S$ to 0, i.e.,

$$\partial L(S)/\partial S = 2SYD_{S1}Y^T - 2CX_g D_{S1}Y^T + 2\beta SYD_{S2}Y^T$$
$$- 2\beta YD_{S2}Y^T + 4\gamma LSYY^T = 0 \tag{24}$$

where $D_{S1} \in \mathcal{R}^{N \times N}$ and $D_{S2} \in \mathcal{R}^{N \times N}$ are diagonal matrices, and $D_{S1,ii} = 1/(2\|(SY - CX_g)_i^T\|)$, $D_{S2,ii} = 1/(2\|(Y - SY)_i^T\|)$. ($A_i^T$ represents the $i$th row in $A^T$.)

Then, Eq. (24) can be re-expressed as

$$(2\gamma L)S + S(YD_{S1}Y^T(YY^T)^{-1} + \beta YD_{S2}Y^T(YY^T)^{-1})$$
$$= CX_g D_{S1}Y^T(YY^T)^{-1} + \beta YD_{S2}Y^T(YY^T)^{-1} \tag{25}$$

Eq. (25) is a classical optimization problem, i.e., the Sylvester equation, which has been studied extensively [43, 45, 59, 60, 61] and widely used [44, 62, 63].

In general, for the Sylvester equation $AW + WB = Z$ ($A \in \mathcal{R}^{m \times m}$, $B \in \mathcal{R}^{n \times n}$, $Z \in \mathcal{R}^{m \times n}$, $W \in \mathcal{R}^{m \times n}$), the matrix $W$ is the variable to be solved. The specific solution to $W$ is given by:

$$W(:) = (I_1 \otimes A + B^T \otimes I_2)^{-1} Z(:) \tag{26}$$

where $I_1 \in \mathcal{R}^{n \times n}$ and $I_2 \in \mathcal{R}^{m \times m}$ are identity matrices, $\otimes$ is the Kronecker tensor product, $Z(:) \in \mathcal{R}^{mn \times 1}$ and $W(:) \in \mathcal{R}^{mn \times 1}$ denote that the matrices $Z$ and $W$ are single column vectors. $W(:)$ can be reshaped to $W^* \in \mathcal{R}^{m \times n}$, which is the solution of $AW + WB = Z$. For simplicity, the solution $W^*$ is denoted as $W^* = sylvester(A, B, Z)$.

Therefore, the solution of Eq. (25) is

$$S^* = sylvester(2\gamma L, Y(D_{S1} + \beta D_{S2})Y^T(YY^T)^{-1},$$
$$(CX_g D_{S1} + \beta YD_{S2})Y^T(YY^T)^{-1}) \tag{27}$$

#### 2) $C$-Subproblem

By fixing $S$, the $C$-subproblem can be expressed as:

$$C^* = argmin_C \|(SY - CX_g)^T\|_{2,1} + \alpha\|C\|_F^2 +$$
$$2\gamma\text{Tr}(Y^T S^T L S Y) \tag{28}$$

In Eq. (28), the Lagrange function for $C$ is

$$L(C) = \|(SY - CX_g)^T\|_{2,1} + \alpha\|C\|_F^2 + 2\gamma\text{Tr}(Y^T S^T L S Y)$$
$$= \|(SY - CX_g)^T\|_{2,1} + \alpha\|C\|_F^2 + 2\gamma\text{Tr}(Y^T S^T (D - R) S Y)$$
$$= \|(SY - CX_g)^T\|_{2,1} + \alpha\|C\|_F^2 + 2\gamma\text{Tr}(Y^T S^T (CC^T \mathbf{1}\mathbf{1}^T \circ$$
$$I_3 - CC^T) S Y) \tag{29}$$

where $\mathbf{1} \in \mathcal{R}^{L \times 1}$ is a column vector with all elements equal to one. The symbol ($\circ$) represents the Hadamard product. $I_3 \in \mathcal{R}^{L \times L}$ is the identity matrix.

Set the derivative of Eq. (29) with respect to $C$ to 0, i.e.,

$$\partial L(C)/\partial C = 2CX_g D_C X_g^T - 2SYD_C X_g^T + 2\alpha C +$$
$$2\gamma(((SYY^T S^T) \circ I_3)^T \mathbf{1}\mathbf{1}^T C + \mathbf{1}\mathbf{1}^T((SYY^T S^T) \circ I_3)C -$$
$$2SYY^T S^T C) = 0 \tag{30}$$

where $D_C \in \mathcal{R}^{N \times N}$ is a diagonal matrix, and $D_{C,ii} = 1/(2\|(SY - CX_g)_i^T\|)$. ($A_i^T$ is the $i$th row of $A^T$.)

Eq. (30) is also a Sylvester equation. Therefore, we can solve $C$ as follows:

$$C^* = sylvester(\alpha I_3 + \gamma((SYY^T S^T) \circ I_3)^T \mathbf{1}\mathbf{1}^T +$$
$$\gamma \mathbf{1}\mathbf{1}^T((SYY^T S^T) \circ I_3) - 2\gamma SYY^T S^T, X_g D_C X_g^T, SYD_C X_g^T) \tag{31}$$

### D. Algorithm Description and Prediction

The procedure of the proposed R-MLTSK-FS is described in Algorithm I.

When the optimal $S^*$ and $C^*$ are obtained, the prediction output of the test instance $x'$ (i.e., $y' = [y_1', ..., y_L']^T$) can be formulated as follows:

$$y' = \varphi_\tau(C^* x_g') \tag{32}$$

where $x_g'$ is the fuzzy feature representation of $x'$ through fuzzy rules. It can be obtained from Eqs. (2)-(4) and (7)-(9). To obtain the final output (i.e., binary label vector), a threshold function $\varphi_\tau(\cdot)$ is introduced to convert the continuous output to the discrete output, and $\tau$ is the threshold. Therefore, for the $l$th label $y_l'$ ($1 \leq l \leq L$) in $y'$, its definition is

$$y_l' = \varphi_\tau((C^* x_g')_l) = \begin{cases} 1, & if\ (C^* x_g')_l \geq \tau \\ 0, & otherwise \end{cases} \tag{33}$$

where $(C^*x'_g)_l$ is the $l$th element in $(C^*x'_g)$. The value of $\tau$ can be optimized by cross-validation. In this paper, we set it to the fixed value of 0.5.

In addition, we use a toy example to illustrate the computation process of the proposed R-MLTSK-FS. Due to the space limitation, the details are shown in the Part A of *Supplementary Materials*.

---

**Algorithm I** R-MLTSK-FS

**Input:** Input matrix $X \in \mathcal{R}^{D \times N}$, output matrix $Y \in \mathcal{R}^{L \times N}$, rule number $K$, trade-off parameters $\alpha$, $\beta$ and $\gamma$, minimum loss margin $Min$, iteration number $T$.

**Procedure:**
1: Transform $X$ into $X_g$ using Eqs. (2)-(4) and (7)-(9).
2: **Initialize:**
  $S^{(0)} = \mathbf{1}_{L \times L}$, $C^{(0)} = (1/L)\mathbf{1}_{L \times K(D+1)}$, $D = \mathbf{0}_{L \times L}$, $loss^{(0)} \leftarrow 0$, $t \leftarrow 1$.
3: **While** $t \le T$ **do**
4: $\quad D_{C,ii} = 1/(2\|(S^{(0)}Y - C^{(0)}X_g)_i^T\|)$;
5: $\quad T1 \leftarrow \alpha I_3 + \gamma((S^{(0)}YY^T(S^{(0)})^T) \circ I_3)^T \mathbf{1}\mathbf{1}^T + \gamma\mathbf{1}\mathbf{1}^T((S^{(0)}YY^T(S^{(0)})^T) \circ I_3) - 2\gamma S^{(0)}YY^T(S^{(0)})^T$;
6: $\quad T2 \leftarrow X_g D_C X_g^T$;
7: $\quad T3 \leftarrow S^{(0)}YD_C X_g^T$;
8: $\quad C^{(t)} \leftarrow sylvester(T1, T2, T3)$;
9: $\quad D_{S1,ii} = 1/(2\|(S^{(0)}Y - C^{(0)}X_g)_i^T\|)$;
10: $\quad D_{S2,ii} = 1/(2\|(Y - S^{(0)}Y)_i^T\|)$;
11: $\quad R \leftarrow C^{(0)}(C^{(0)})^T$;
12: $\quad D_{ii} \leftarrow \sum_{j=1}^L R_{ij}$;
13: $\quad L = D - R$;
14: $\quad T4 \leftarrow 2\gamma L$;
15: $\quad T5 \leftarrow Y(D_{S1} + \beta D_{S2})Y^T(YY^T)^{-1}$;
16: $\quad T6 \leftarrow (C^{(0)}X_g D_{S1} + \beta YD_{S2})Y^T(YY^T)^{-1}$;
17: $\quad S^{(t)} \leftarrow sylvester(T4, T5, T6)$;
18: $\quad C^{(0)} \leftarrow C^{(t)}$;
19: $\quad S^{(0)} \leftarrow S^{(t)}$;
20: $\quad loss^{(t)} \leftarrow \|(S^{(t)}Y - C^{(t)}X_g)^T\|_{2,1} + \alpha\|C^{(t)}\|_F^2 + \beta\|(Y - S^{(t)}Y)^T\|_{2,1} + 2\gamma Tr(Y^T(S^{(t)})^T LS^{(t)}Y)$;
21: **If** $|loss^{(t)} - loss^{(0)}| \le Min$ or $loss^{(t)} \le 0$ **do**
22: $\quad$ break, to End While;
23: **Else**
24: $\quad loss^{(0)} \leftarrow loss^{(t)}$, $t \leftarrow t + 1$;
25: **End If**
26: **End While**
**Output:** $S$, $C$.

---

### E. Computational Complexity Analysis

The computational complexity of R-MLTSK-FS is analyzed according to the steps in Algorithm I, which is expressed using the big-O notation. For step 1, the computational complexity of transforming $X$ into $X_g$ is $O(2NKD + 2NK)$. For step 2, the complexity of initialization is $O(1)$. The computational complexity of step 4 is $O(L^2N + LNK(D + 1))$. For the step 5, the computational complexity of $T1$ is $O(2L^2N + L^3 + 2L^2)$. For step 6, the computational complexity of $T2$ is $O(N^2K(D + 1) + NK^2(D + 1)^2)$. For step 7, the computational complexity of calculating $T3$ is $O(L^2N + LN^2 + LNK(D + 1))$. The computational complexity of step 8 is $O(3L^4)$. For step 9, the complexity of calculating $D_{S1}$ is $O(L^2N + LNK(D + 1))$. For step 10, the complexity of $D_{S2}$ is $O(L^2N)$. The complexity of step 11 is $O(L^2K(D + 1))$. The complexity of steps 12-14 is $O(1)$. For step 15, the complexity of $T5$ is $O(LN^2 + L^2N + L^3)$. The complexity of step 16 is $O(LNK(D + 1) + LN^2 + L^2N + L^3)$. For step 17, the complexity is $O(3L^2K^2(D + 1)^2)$. For steps 18 and 19, the complexity is $O(1)$ respectively. For step 20, the computational complexity is $O(LNK(D + 1) + LN^2 + 2L^2N)$. For steps 21-24, the complexity is $O(1)$. Hence, the overall complexity of the whole algorithm is dominated by steps 6 and 20. Let $a = \max(L, D, K)$, $b = \max(N, K(D + 1))$. In general, $a \ll b$. Therefore, the maximum computational complexity of R-MLTSK-FS is $O(2b(a^2 + ab + b^2))$.

## IV. EXPERIMENTAL ANALYSIS

Extensive experiments are conducted to fully assess the effectiveness of R-MLTSK-FS, including classification performance evaluation, robustness analysis, effectiveness analysis of soft label learning and correlation enhancement learning, cross-validation based parameter analysis, convergence analysis, statistical analysis, running time analysis and interpretability analysis. The datasets, evaluation metrics and the settings used in the experiments are described below.

### A. Datasets

We adopt 10 benchmark multilabel datasets to evaluate the performance of R-MLTSK-FS. Table I shows the details of these datasets, where #Ins, #Fea and #Lab denote the instance number, feature dimension, and label space dimension respectively. These datasets are available from the Github[1].

TABLE I
STATISTICS OF DATASETS

| Dataset | #Ins | #Fea | #Lab |
|---|---|---|---|
| Arts | 5000 | 462 | 26 |
| Birds | 645 | 260 | 19 |
| CAL500 | 502 | 68 | 174 |
| Corel5k | 5000 | 499 | 374 |
| Flags | 194 | 19 | 7 |
| Genbase | 662 | 1185 | 27 |
| Medical | 978 | 1449 | 45 |
| Mirflickr | 25000 | 150 | 24 |
| Recreation | 5000 | 606 | 22 |
| Science | 5000 | 743 | 40 |

### B. Evaluation Metrics

Let $\{(\tilde{x}_i, \tilde{y}_i) | 1 \le i \le N_t\}$ be a test set with $N_t$ samples, $\hat{y}_i$ be the predicted labels of $\tilde{x}_i$, $f(\tilde{x}_i, l)$ be the continuous output predicted by the multilabel method for the instance $\tilde{x}_i$ on the $l$th label. The ranking function $rank(\tilde{x}_i, l)$ is obtained according to $f(\tilde{x}_i, l)$. If $f(\tilde{x}_i, l) > f(\tilde{x}_i, l')$, then $rank(\tilde{x}_i, l) < rank(\tilde{x}_i, l')$. Let $L_{x_i}$ be the label set related to $\tilde{x}_i$, and $\overline{L}_{x_i}$ is the complement of $L_{x_i}$. Based on the settings, the four metrics below, which are commonly used in multilabel learning, are employed in the experiments [46].

(1) Average Precision (AP): It is the average proportion of the related labels of an instance that are ranked lower than a given label $l$. The larger the value of AP, the better the classification performance.

---

[1] https://github.com/ZesenChen/multi-label-dataset and https://github.com/KKimura360/MLC_toolbox/tree/master/dataset/matfile

$$\text{AP} = \frac{1}{N_t}\sum_{i=1}^{N_t}\frac{1}{|L_{x_i}|}\sum_{l\in L_{x_i}}\frac{\left|\{l'\in L_{x_i}|f(\widetilde{x}_i,l')\geq f(\widetilde{x}_i,l)\}\right|}{rank(\widetilde{x}_i,l)} \quad (34)$$

(2) Hamming Loss (HL): It is the average proportion of an instance that is predicted incorrectly. The smaller the value of HL, the better the classification performance.

$$\text{HL} = \frac{1}{N_t}\sum_{i=1}^{N_t}\frac{|\widetilde{y}_i\oplus\widehat{y}_i|}{L} \quad (35)$$

where $\oplus$ is the XOR operation.

(3) Ranking Loss (RL): It is the proportion of the related labels that are ranked higher than the unrelated labels. The smaller the value of RL, the better the classification performance.

$$\text{RL} = \frac{1}{N_t}\sum_{i=1}^{N_t}\frac{\left|\{(l,l')|f(\widetilde{x}_i,l)\leq f(\widetilde{x}_i,l'),(l,l')\in L_{x_i}\times \overline{L}_{x_i}\}\right|}{|L_{x_i}||\overline{L}_{x_i}|} \quad (36)$$

(4) Coverage (CV): It is the average number of times that all related labels of an instance are found. The smaller the value of CV, the better the classification performance.

$$\text{CV} = \frac{1}{N_t}\sum_{i=1}^{N_t}\max_{l\in L_{x_i}} rank(\widetilde{x}_i,l) - 1 \quad (37)$$

### C. Experimental Settings

In this paper, we employ eleven methods for comparison, including binary relevance (BR) [47], multilabel $k$-nearest neighbor (ML$k$NN) [48], meta-label-specific features (MLSF) [49], ML-TSK FS [17], classifier chains (CC) [50], random $k$-labelsets (RA$k$EL) [51], correlated logistic models (CorrLog) [52], hybrid noise-oriented multilabel learning (HNOML) [53], multilabel learning with local similarity of samples (ML-LSS) [64], shared weight matrix with low-rank and sparse regularization for multilabel learning (2SML) [65] and robust label and feature space co-learning (RLFSCL) [66]. These methods and the settings of the parameters using grid search are described in Table II. We adopt five-fold cross-validation strategy to evaluate the performance.

TABLE II
DESCRIPTION OF METHODS

| Methods | Description | Parameter Setting |
|---|---|---|
| BR | This method is a first-order method. To improve the robustness, it introduces $\varepsilon$-insensitive learning (a fuzzy method) by solving a system of linear inequalities ($\varepsilon$LSSLI) [54] as the binary classifier. The maximum computational complexity of this method is $O(ab^2(a+6b))$, where $a=\max(L,D,M)$ and $b=\max(N,M(D+1))$. | $C = 2.\wedge(-5:1:5)$, $M = \{2,3,4,5,6,7,8,9\}$. |
| ML$k$NN | This method is a first-order method that predicts a new instance by maximizing the posterior probability of each label. The number of nearest neighbors affects the robustness of the model to some extent. The maximum computational complexity of this method is $O(ab(a+b))$, where $a=\max(L,D,K)$ and $b=N$. | $K = \{1,3,5,7,9,11,13\}$, $s = \{0.01, 0.03, 0.05, 0.07, 0.09\}$. |
| MLSF | This method is a second-order method. It improves the performance through meta-label learning and specific feature selection. The maximum computational complexity of this method is $O(a^2b + 3a^3 + a^2b^2 + a^3b)$, where $a=\max(L,D,k)$ and $b=N$. | $k = \{2,4,6,8\}$, $\varepsilon = \{0.1,1,10\}$, $\alpha = \{0.1,0.5,0.9\}$, $\gamma = \{0.1,1,10\}$. |
| ML-TSK FS | This method is a second-order method that uses the correlation between any two labels to improve performance. To realize a transparent model, it uses fuzzy rules to model the inference relationship between features and labels. This method does not consider the influence of label noise. The maximum computational complexity of this method is $O(ab^2(a+2b))$, where $a=\max(L,K)$ and $b=\max(N,K(D+1))$. | $K = \{2,3,4,5\}$, $\alpha = \{0.01,0.1,1,10,100\}$, $\beta = \{0.01,0.1,1,10,100\}$. |
| CC | This method is a high-order method which takes the prediction result of the previous label into account in the feature space predicting the next label. The $\varepsilon$-insensitive learning (a fuzzy method) by solving a system of linear inequalities ($\varepsilon$LSSLI) [54] is used as the binary classifier to improve robustness. The maximum computational complexity of this method is $O(a^2 + b^2(a+6b) + 2ab(2a+4ab+a^3))$, where $a=\max(L,D+1,M)$ and $b=\max(N,M(D+L))$. | $C = 2.\wedge(-5:1:5)$, $M = \{2,3,4,5,6,7,8,9\}$. |
| RA$k$EL | This method is a high-order method. In this method, the label space is randomly divided into multiple label subspaces, and the prediction result of a label is associated with other labels in the subspace. The maximum computational complexity of this method is $O(2L\cdot F_m(N,D,2^k))$, where $F_m()$ is the complexity of multi-classification classifier. | $k = N./(12:-2:2)$ ($N$ is the instance number), $\alpha = \{0.1, 0.3, 0.5, 0.7, 0.9\}$. |
| CorrLog | This method is a high-order method. It achieves robustness by constructing the association between a label and all other labels. The maximum computational complexity of this method is $O(2a^3b)$, where $a=\max(L,D)$ and $b=N$. | $rho1 = \{0.001, 0.003, 0.005, 0.007, 0.009, 0.01, 0.03, 0.05, 0.07, 0.09, 0.1, 0.3, 0.5, 0.7, 0.9\}$, $rho2 = \{0.001, 0.005, 0.01, 0.05, 0.1, 0.5\}$. |
| HNOML | This method is a high-order method. It designs a label enrichment matrix to improve the robustness. The maximum computational complexity of this method is $O(3a^3 + 2ab^2 + 4a^2b)$, where $a=\max(L,D)$ and $b=N$. | $\alpha = \{0.01,0.1,1,10\}$, $\beta = \{0.01,0.1,1,10,100\}$, $\gamma = \{0.01,0.1,1,10\}$. |
| ML-LSS | This method is a second-order method. It builds a new space where the similarity between labels is utilized to strengthen the similarity between samples, thus improving the classification performance. The maximum computational complexity of this method is $O(a^2 + b^2 + 2ab + 2a^2b + 2ab^2)$, where $a=\max(L,D)$ and $b=N$. | $\lambda_1 = 2.\wedge(-5:1:6)$, $\lambda_2 = 2.\wedge(-5:1:6)$. |
| 2SML | This method is a second-order method that adopts cosine similarity to measure the correlation between labels. The maximum computational complexity of this method is $O(2a(3ab+2a^2+b^2))$, where $a=\max(L,D)$ and $b=N$. | $\lambda_1 = 10\wedge(-4:1:3)$, $\lambda_2 = 10\wedge(-4:1:3)$, $\lambda_3 = 10\wedge(-4:1:3)$. |
| RLFSCL | This method is a high-order method. It achieves label correlation learning and enhances robustness by constructing a low rank label space from traditional regression learning. The maximum computational complexity of this method is $O(5a^2(a+b))$, where $a=\max(L,D)$ and $b=N$. | $\lambda = 10.\wedge(-2:1:2)$, $\mu = 10.\wedge(3:1:5)$. |
| **R-MLTSK-FS (ours)** | The method proposed in this paper. It is a second-order method and achieves transparency and robustness against label noise through fuzzy rules, correlation enhancement learning, soft multilabel loss function construction, and soft label learning. | $\alpha = \{0.001,0.005,0.01,0.05,0.1,0.5,1,5,10,50,100\}$, $\beta = \{0.001,0.005,0.01,0.05,0.1,0.5,1,5,10,50,100\}$, $\gamma = \{0.001,0.005,0.01,0.05,0.1,0.5,1,5,10,50,100\}$, $k = \{2,3\}$. |

maximum computational complexity of this method is $O(2b(a^2 + ab + b^2))$, where $a = \max(L, D, K)$ and $b = \max(N, K(D+1))$.

TABLE III
MEAN (SD) OF THE METRICS OF THE MULTILABEL CLASSIFICATION METHODS

| Datasets | Metrics | BR | MLkNN | MLSF | ML-TSK FS | CC | RAkEL | CorrLog | HNOML | ML-LSS | 2SML | RLF-SCL | R-MLTSK-FS |
|---|---|---|---|---|---|---|---|---|---|---|---|---|---|
| Arts | AP | 0.6270 (0.0076) | 0.5454 (0.0082) | 0.4977 (0.0859) | 0.6207 (0.0141) | 0.6164 (0.0084) | 0.2682 (0.0285) | 0.3646 (0.0482) | 0.6090 (0.0082) | 0.6273 (0.0040) | 0.6247 (0.0129) | 0.5810 (0.0123) | **0.6289 (0.0130)** |
| | HL | 0.0902 (0.0050) | 0.0629 (0.0007) | 0.0604 (0.0022) | **0.0529 (0.0019)** | 0.1025 (0.0011) | 0.1950 (0.0092) | 0.0597 (0.0018) | 0.0573 (0.0009) | 0.5056 (0.0025) | 0.0558 (0.0021) | 0.0561 (0.0012) | 0.0546 (0.0017) |
| | RL | 0.1266 (0.0042) | 0.1396 (0.0028) | 0.1257 (0.0309) | 0.1161 (0.0039) | 0.1300 (0.0069) | 0.4123 (0.0325) | 0.3865 (0.0878) | 0.1509 (0.0052) | 0.1163 (0.0065) | 0.1185 (0.0056) | 0.1619 (0.0066) | **0.1118 (0.0075)** |
| | CV | 0.1965 (0.0053) | 0.1981 (0.0036) | 0.3047 (0.0663) | 0.1807 (0.0083) | 0.2054 (0.0082) | 0.8363 (0.0369) | 0.4724 (0.0694) | 0.2371 (0.0045) | 0.1786 (0.0112) | 0.1847 (0.0093) | 0.4746 (0.0068) | **0.1720 (0.0073)** |
| Birds | AP | 0.3422 (0.0340) | 0.2303 (0.0185) | 0.2712 (0.0203) | 0.3438 (0.0347) | 0.3360 (0.0174) | 0.3591 (0.0319) | 0.2124 (0.0230) | 0.3352 (0.0325) | 0.3230 (0.0436) | 0.2769 (0.0439) | 0.2965 (0.0276) | **0.3694 (0.0354)** |
| | HL | 0.0556 (0.0022) | 0.0551 (0.0058) | 0.0648 (0.0027) | 0.0514 (0.0038) | 0.0545 (0.0033) | 0.0446 (0.0032) | 0.0451 (0.0027) | 0.0515 (0.0065) | 0.0456 (0.0059) | 0.3289 (0.0520) | 0.0627 (0.0054) | **0.0430 (0.0063)** |
| | RL | 0.0983 (0.0230) | 0.1565 (0.0127) | 0.0807 (0.0205) | 0.0863 (0.0221) | 0.1097 (0.0055) | 0.6509 (0.0634) | 0.1611 (0.0067) | 0.0968 (0.0215) | 0.1543 (0.0321) | 0.4480 (0.0415) | 0.2200 (0.0281) | **0.0710 (0.0124)** |
| | CV | 0.1311 (0.0151) | 0.1887 (0.0203) | 0.1699 (0.0495) | 0.1132 (0.0315) | 0.1445 (0.0094) | 0.7032 (0.0364) | 0.1939 (0.0141) | 0.1179 (0.0188) | 0.1135 (0.0298) | 0.5403 (0.0320) | 0.2944 (0.0239) | **0.0957 (0.0193)** |
| CAL500 | AP | 0.5048 (0.0055) | 0.4965 (0.0037) | 0.4906 (0.0119) | 0.5075 (0.0104) | 0.4541 (0.0088) | 0.2150 (0.0047) | 0.3108 (0.0171) | 0.4314 (0.1844) | 0.5095 (0.0088) | 0.4460 (0.0079) | 0.5069 (0.0059) | **0.5153 (0.0152)** |
| | HL | 0.1447 (0.0034) | 0.1371 (0.0031) | 0.1368 (0.0027) | 0.1368 (0.0027) | 0.1442 (0.0026) | 0.1363 (0.0036) | 0.1371 (0.0046) | 0.1411 (0.0072) | 0.1363 (0.0029) | 0.1461 (0.0011) | 0.1372 (0.0037) | **0.1358 (0.0034)** |
| | RL | 0.1879 (0.0058) | 0.1822 (0.0043) | 0.1780 (0.0053) | 0.1763 (0.0035) | 0.2515 (0.0085) | 0.6145 (0.0161) | 0.6750 (0.1145) | **0.1423 (0.0797)** | 0.1767 (0.0032) | 0.2105 (0.0046) | 0.1858 (0.0017) | 0.1744 (0.0012) |
| | CV | 0.7656 (0.0132) | 0.7583 (0.0122) | 0.7600 (0.0132) | 0.7380 (0.0091) | 0.9085 (0.0105) | 0.7835 (0.0264) | 0.8722 (0.0119) | 0.7669 (0.0579) | 0.7417 (0.0102) | 0.8640 (0.0094) | 0.9729 (0.0026) | **0.7348 (0.0278)** |
| Corel5k | AP | 0.3044 (0.0068) | 0.2561 (0.0077) | 0.2134 (0.0178) | 0.3064 (0.0003) | 0.2639 (0.0061) | 0.0652 (0.0032) | 0.2079 (0.0085) | 0.2884 (0.0105) | 0.3051 (0.0089) | 0.2843 (0.0097) | 0.3058 (0.0082) | **0.3070 (0.0070)** |
| | HL | 0.0094 (0.0001) | 0.0094 (0.0001) | 0.0094 (0.0001) | 0.0094 (0.0003) | 0.0094 (0.0001) | 0.0197 (0.0002) | 0.0094 (0.0003) | 0.0111 (0.0006) | 0.0094 (0.0004) | 0.0097 (0.0007) | 0.0094 (0.0001) | **0.0094 (0.0001)** |
| | RL | 0.1649 (0.0044) | 0.1313 (0.0040) | 0.2591 (0.0290) | 0.1294 (0.0047) | 0.1784 (0.0068) | 0.5564 (0.0279) | 0.1432 (0.0032) | 0.1119 (0.2279) | 0.1265 (0.0019) | 0.2110 (0.0047) | 0.1140 (0.0045) | **0.1092 (0.0028)** |
| | CV | 0.3852 (0.0045) | 0.3023 (0.0059) | 0.6994 (0.0983) | 0.3018 (0.0108) | 0.4288 (0.0108) | 0.5552 (0.0167) | 0.3207 (0.0101) | 0.3678 (0.0092) | 0.2947 (0.0046) | 0.4691 (0.0069) | 0.9328 (0.0030) | **0.2600 (0.0090)** |
| Flags | AP | 0.8101 (0.0316) | 0.8020 (0.0415) | 0.8163 (0.0226) | 0.8176 (0.0118) | 0.8076 (0.0413) | 0.6581 (0.0544) | 0.7704 (0.0180) | 0.8080 (0.0110) | 0.8182 (0.0231) | 0.6921 (0.0292) | 0.8148 (0.0145) | **0.8209 (0.0391)** |
| | HL | 0.2796 (0.0216) | 0.3275 (0.0272) | 0.2768 (0.0155) | 0.2649 (0.0254) | 0.2711 (0.0307) | 0.2755 (0.0323) | 0.2856 (0.0258) | 0.2711 (0.0124) | **0.2511 (0.0093)** | 0.4358 (0.0160) | 0.2615 (0.0309) | 0.2647 (0.0438) |
| | RL | 0.2155 (0.0341) | 0.2443 (0.0374) | **0.1374 (0.0066)** | 0.2132 (0.0173) | 0.2340 (0.0495) | 0.6030 (0.0419) | 0.3566 (0.0408) | 0.2178 (0.0159) | 0.2088 (0.0352) | 0.3993 (0.0307) | 0.2096 (0.0236) | 0.2054 (0.0345) |
| | CV | 0.5523 (0.0159) | 0.5626 (0.0198) | 0.5524 (0.0206) | **0.5232 (0.0127)** | 0.5553 (0.0123) | 0.8903 (0.0252) | 0.5486 (0.0150) | 0.5431 (0.0341) | 0.5331 (0.0071) | 0.6753 (0.0214) | 0.6444 (0.0193) | 0.5318 (0.0276) |
| Genbase | AP | 0.9922 (0.0067) | 0.9910 (0.0043) | 0.9913 (0.0051) | 0.9968 (0.0027) | 0.9802 (0.0181) | 0.7784 (0.0697) | 0.9717 (0.0097) | 0.9941 (0.0050) | 0.9967 (0.0044) | 0.9955 (0.0040) | 0.9886 (0.0099) | **0.9977 (0.0031)** |
| | HL | 0.0011 (0.0006) | 0.0016 (0.0005) | 0.0044 (0.0016) | 0.0015 (0.0017) | 0.0095 (0.0033) | 0.0022 (0.0012) | 0.0022 (0.0007) | 0.0020 (0.0015) | 0.0010 (0.0002) | 0.0010 (0.0003) | 0.0019 (0.0001) | **0.0010 (0.0012)** |
| | RL | 0.0035 (0.0049) | 0.0061 (0.0040) | 0.0038 (0.0026) | 0.0011 (0.0009) | 0.0087 (0.0081) | 0.0242 (0.0184) | 0.0355 (0.0095) | 0.0006 (0.0007) | 0.0012 (0.0013) | 0.0025 (0.0025) | 0.0121 (0.0122) | **0.0006 (0.0005)** |
| | CV | 0.0150 (0.0061) | 0.0192 (0.0073) | 0.0195 (0.0073) | 0.0105 (0.0042) | 0.0244 (0.0154) | 0.0588 (0.0159) | 0.0407 (0.0063) | 0.0126 (0.0046) | 0.0113 (0.0040) | 0.0136 (0.0045) | 0.0239 (0.0122) | **0.0102 (0.0021)** |
| Medical | AP | 0.8755 (0.0266) | 0.8067 (0.0128) | 0.8272 (0.0250) | 0.8959 (0.0143) | 0.8765 (0.0307) | 0.4443 (0.0219) | 0.7562 (0.0181) | 0.8761 (0.0495) | 0.8822 (0.0091) | **0.9059 (0.0107)** | 0.8386 (0.0181) | 0.8822 (0.0150) |
| | HL | 0.0142 (0.0018) | 0.0156 (0.0004) | 0.0131 (0.0012) | 0.0107 (0.0006) | 0.0125 (0.0014) | 0.0109 (0.0008) | 0.0113 (0.0007) | 0.0213 (0.0085) | 0.0114 (0.0015) | 0.0112 (0.0014) | 0.0137 (0.0014) | **0.0105 (0.0019)** |
| | RL | 0.0274 (0.0147) | 0.0430 (0.0061) | 0.0273 (0.0038) | 0.0371 (0.0136) | 0.0311 (0.0175) | 0.1079 (0.0250) | 0.2742 (0.0258) | 0.0232 (0.0320) | 0.0300 (0.0051) | **0.0167 (0.0049)** | 0.0470 (0.0121) | 0.0197 (0.0039) |
| | CV | 0.0415 (0.0186) | 0.0629 (0.0056) | 0.0717 (0.0082) | 0.0363 (0.0068) | 0.0453 (0.0226) | 0.1394 (0.0304) | 0.1969 (0.0280) | 0.0357 (0.0217) | 0.0460 (0.0130) | **0.0331 (0.0053)** | 0.1700 (0.0313) | 0.0308 (0.0105) |
| Mirflickr | AP | 0.4540 (0.0421) | 0.5096 (0.0028) | 0.2906 (0.0156) | 0.5239 (0.0045) | 0.4703 (0.0019) | 0.2216 (0.0030) | 0.4779 (0.0085) | 0.5121 (0.0084) | 0.5222 (0.0022) | 0.5201 (0.0030) | 0.5053 (0.0027) | **0.5246 (0.0015)** |
| | HL | 0.1528 (0.0122) | 0.1533 (0.0006) | 0.1543 (0.0010) | 0.1521 (0.0005) | 0.1588 (0.0010) | 0.2122 (0.0030) | 0.1548 (0.0005) | 0.1523 (0.0022) | 0.1525 (0.0010) | 0.1557 (0.0012) | 0.1556 (0.0015) | **0.1521 (0.0004)** |
| | RL | 0.3218 (0.0419) | 0.2050 (0.0027) | 0.2616 (0.0012) | 0.1946 (0.0015) | 0.2444 (0.0015) | 0.5694 (0.0087) | 0.2146 (0.0028) | 0.2106 (0.0097) | 0.2054 (0.0010) | 0.2066 (0.0013) | 0.2276 (0.0012) | **0.1929 (0.0012)** |
| | CV | 0.6120 (0.0327) | 0.4395 (0.0045) | 0.4703 (0.0082) | 0.4190 (0.0031) | 0.5314 (0.0037) | 0.9937 (0.0021) | 0.4495 (0.0041) | 0.4434 (0.0043) | 0.4329 (0.0017) | 0.4437 (0.0032) | 0.6189 (0.0016) | **0.4182 (0.0051)** |
| Recreation | AP | 0.6363 (0.0151) | 0.5333 (0.0092) | 0.4817 (0.0426) | 0.6362 (0.0061) | 0.6286 (0.0152) | 0.2922 (0.0193) | 0.2104 (0.0247) | 0.6062 (0.0076) | 0.6287 (0.0092) | 0.6297 (0.0067) | 0.5901 (0.0091) | **0.6366 (0.0058)** |
| | HL | 0.0905 (0.0014) | 0.0647 (0.0012) | 0.0637 (0.0014) | 0.0592 (0.0012) | 0.0998 (0.0019) | 0.2923 (0.0148) | 0.0583 (0.0010) | 0.0563 (0.0021) | 0.0558 (0.0012) | 0.0556 (0.0010) | 0.0556 (0.0019) | **0.0553 (0.0017)** |
| | RL | 0.1391 (0.0082) | 0.1640 (0.0011) | 0.1408 (0.0410) | 0.1297 (0.0020) | 0.1400 (0.0083) | 0.4073 (0.0155) | 0.4839 (0.0119) | 0.1989 (0.0061) | 0.1255 (0.0029) | 0.1267 (0.0054) | 0.1841 (0.0092) | **0.1246 (0.0058)** |
| | CV | 0.1877 (0.0117) | 0.2035 (0.0040) | 0.3076 (0.0867) | 0.1697 (0.0043) | 0.1906 (0.0125) | 0.8912 (0.0206) | 0.4554 (0.0240) | 0.2545 (0.0113) | 0.1680 (0.0076) | 0.1739 (0.0079) | 0.5105 (0.0077) | **0.1675 (0.0054)** |
| Science | AP | 0.5983 (0.0132) | 0.5134 (0.0119) | 0.4461 (0.0063) | 0.5978 (0.0217) | 0.5861 (0.0125) | 0.2333 (0.0115) | 0.2492 (0.0106) | 0.5737 (0.0144) | 0.5918 (0.0058) | 0.5963 (0.0074) | 0.5567 (0.0058) | **0.5984 (0.0051)** |
| | HL | 0.0526 (0.0007) | 0.0363 (0.0006) | 0.0343 (0.0011) | 0.0329 (0.0004) | 0.0603 (0.0009) | 0.1288 (0.0087) | 0.0370 (0.0036) | 0.0333 (0.0004) | 0.0330 (0.0010) | 0.0326 (0.0011) | 0.0336 (0.0015) | **0.0324 (0.0009)** |
| | RL | 0.1140 | 0.1211 | 0.0990 | 0.0996 | 0.1128 | 0.3794 | 0.4989 | 0.1867 | 0.0985 | 0.1040 | 0.1438 | **0.0976** |

|    |  (0.0068) | (0.0046) | (0.0143) | (0.0072) | (0.0071) | (0.0352) | (0.1339) | (0.0086) | (0.0042) | (0.0058) | (0.0060) | **(0.0050)** |
|----|-----------|----------|----------|----------|----------|----------|----------|----------|----------|----------|----------|--------------|
| CV | 0.1596    | 0.1574   | 0.1823   | 0.1357   | 0.1620   | 0.7443   | 0.3614   | 0.2434   | 0.1336   | 0.1474   | 0.5376   | **0.1321**   |
|    | (0.0089)  | (0.0050) | (0.0269) | (0.0088) | (0.0093) | (0.0366) | (0.0219) | (0.0061) | (0.0058) | (0.0093) | (0.0088) | **(0.0058)** |

### D. Performance Analysis

#### (1) Classification Performance Evaluation

To verify the effectiveness of R-MLTSK-FS, we compare the R-MLTSK-FS with eleven methods on 10 datasets. Table III shows the experimental results, expressed in terms of the mean and standard deviation (inside brackets) of the AP, HL, RL and CV. For each dataset, the best value of each metric is bold-faced. We can see that the overall performance of R-MLTSK-FS is the best on the four metrics. This is attributable to the three mechanisms introduced.

#### (2) Robustness Analysis

In order to verify the robustness of R-MLTSK-FS against label noise, we introduce label noise to the data and use the AP results as an example to evaluate the performance. Five-fold cross-validation strategy is adopted in the experiment. For each fold, we refer to the approaches in [53, 57, 58] to generate label noise. That is, we randomly select 0%, 10%, 20%, 30% and 40% samples from the training set and then generate label noise for the selected training samples by changing their related labels to unrelated ones, and vice versa. Fig. 2 shows the experimental results, from which the following findings are obtained:

1) On the whole, despite the increase in the amount of noise, the proposed R-MLTSK-FS maintains outstanding classification performance, indicating the effectiveness of the three mechanisms introduced in reducing the influence of label noise.

2) Label noise has different effect on the comparison methods. For example, the performance of ML$k$NN in the presence of label noise is unstable because the robustness of ML$k$NN against noisy labels is affected by the number of nearest neighbors. For RA$k$EL and CorrLog, their performance is unsatisfactory since they ignore label noise in modeling the correlation between labels. For ML-TSK FS and 2SML, their overall robustness is inferior to the proposed method as they also ignore the influence of label noise in model training. For RLFSCL, its performance is mediocre but the robustness against label noise is stable, which indicates that constructing a low rank label space can resist the interference of label noise to a certain extent. For ML-LSS, its robustness is unstable since the influence of label noise is ignored in space transformation.

#### (3) Effectiveness Analysis of Soft Label Learning

To fully evaluate the effectiveness of R-MLTSK-FS in soft label learning, we conduct influence weights ($S$) analysis and ablation experiment analysis for soft label learning in this section. The details are as follows.

*1) Influence Weights $S$ Analysis:* We study the influence weights $S$ with three synthetic multilabel datasets, namely *Independence dataset*, *Equality dataset* and *Union dataset* [55], each containing 1000 samples. For each sample, the feature dimension is 20 and the label dimension is 5. The features in the synthetic datasets are normalized in [0, 1].

Each synthetic dataset has five labels, $\mathcal{Y}_1, \ldots, \mathcal{Y}_5$. For the first four labels, their logical relationships are designed as follows:

*Independence dataset:* The first four labels $\mathcal{Y}_1$, $\mathcal{Y}_2$, $\mathcal{Y}_3$ and $\mathcal{Y}_4$ are independent of each other.

*Equality dataset:* $\mathcal{Y}_1 = \mathcal{Y}_2$ and $\mathcal{Y}_3 = \mathcal{Y}_4$. That is, for a sample $(x_i, y_i)$ ($1 \leq i \leq 1000$), $y_{i1} = y_{i2}$ and $y_{i3} = y_{i4}$.

*Union dataset:* $\mathcal{Y}_1 = \mathcal{Y}_2 \vee \mathcal{Y}_3 \vee \mathcal{Y}_4$. That is, for a sample $(x_i, y_i)$ ($1 \leq i \leq 1000$), if $y_{i2} = 1$ or $y_{i3} = 1$ or $y_{i4} = 1$, then $y_{i1} = 1$, otherwise, $y_{i1} = 0$.

The fifth label is mutually exclusive with the first four labels (i.e., $\mathcal{Y}_5 = (\neg \mathcal{Y}_1) \wedge (\neg \mathcal{Y}_2) \wedge (\neg \mathcal{Y}_3) \wedge (\neg \mathcal{Y}_4)$). Specifically, for a sample $(x_i, y_i)$ ($1 \leq i \leq 1000$), if $y_{i1} = 0$ and $y_{i2} = 0$ and $y_{i3} = 0$ and $y_{i4} = 0$, then $y_{i5} = 1$, otherwise, $y_{i5} = 0$.

The learned influence weights $S$ for each of the three synthetic datasets are shown in Tables IV-VI respectively. The following findings can be obtained from the tables:

1) In Tables IV-VI, since the fifth label is mutually exclusive with the first four labels (i.e., $\mathcal{Y}_5 = (\neg \mathcal{Y}_1) \wedge (\neg \mathcal{Y}_2) \wedge (\neg \mathcal{Y}_3) \wedge (\neg \mathcal{Y}_4)$), reconstruction cannot be achieved with the first four labels. From the results of influence weights in

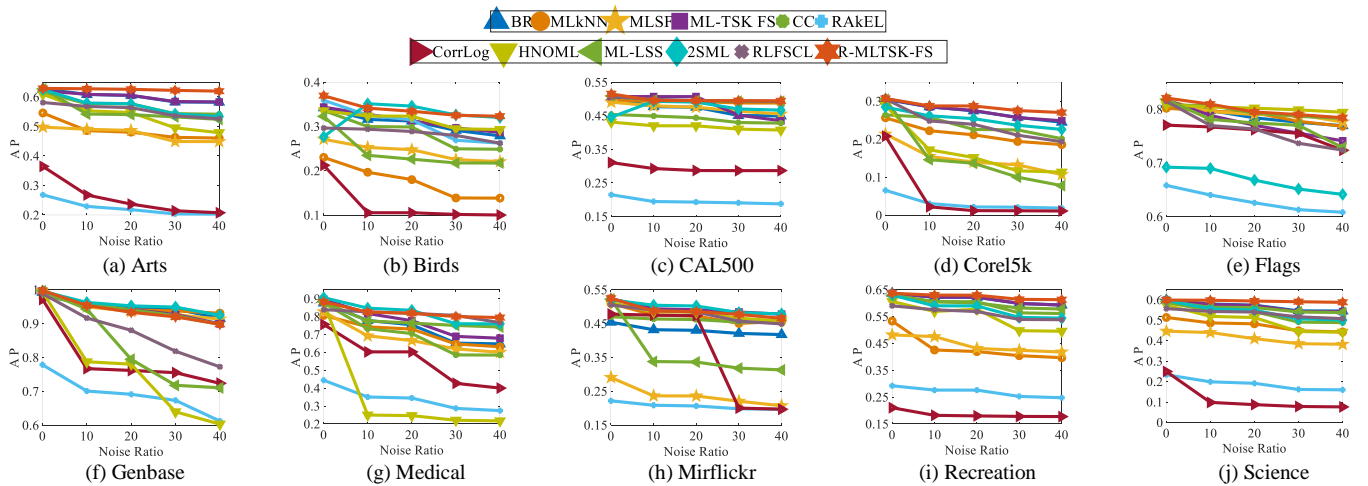

(a) Arts  (b) Birds  (c) CAL500  (d) Corel5k  (e) Flags
(f) Genbase  (g) Medical  (h) Mirflickr  (i) Recreation  (j) Science

Fig. 2 Performance in terms of AP on datasets with label noise. (Noise ratio is defined as the proportion of samples that are randomly selected from the training set and their related (unrelated) labels are changed to unrelated (related) ones. The larger the value of AP, the better the classification performance.)

TABLE IV
INFLUENCE WEIGHTS ($S$) OF ORIGINAL LABELS ON A SOFT LABEL IN INDEPENDENCE DATASET

|  | original label 1 ($\mathcal{Y}_1$) | original label 2 ($\mathcal{Y}_2$) | original label 3 ($\mathcal{Y}_3$) | original label 4 ($\mathcal{Y}_4$) | original label 5 ($\mathcal{Y}_5$) |
| --- | --- | --- | --- | --- | --- |
| soft label 1 ($\mathcal{Y}'_1$) | **0.2016** | 0.0510 | 0.0697 | 0.0462 | **0.0797** |
| soft label 2 ($\mathcal{Y}'_2$) | 0.1409 | **0.3149** | 0.1921 | 0.1552 | **0.2182** |
| soft label 3 ($\mathcal{Y}'_3$) | 0.2447 | 0.2523 | **0.4662** | 0.2628 | **0.3666** |
| soft label 4 ($\mathcal{Y}'_4$) | 0.0031 | 0.0051 | 0.0053 | **0.1191** | 0.0061 |
| soft label 5 ($\mathcal{Y}'_5$) | 0.1179 | 0.1046 | 0.1068 | 0.1281 | **0.2832** |

N.B. $\mathcal{Y}_1$, $\mathcal{Y}_2$, $\mathcal{Y}_3$ and $\mathcal{Y}_4$ are independent. $\mathcal{Y}_5 = (\neg \mathcal{Y}_1) \wedge (\neg \mathcal{Y}_2) \wedge (\neg \mathcal{Y}_3) \wedge (\neg \mathcal{Y}_4)$.

TABLE V
INFLUENCE WEIGHTS ($S$) OF ORIGINAL LABELS ON A SOFT LABEL IN EQUALITY DATASET

|  | original label 1 ($\mathcal{Y}_1$) | original label 2 ($\mathcal{Y}_2$) | original label 3 ($\mathcal{Y}_3$) | original label 4 ($\mathcal{Y}_4$) | original label 5 ($\mathcal{Y}_5$) |
| --- | --- | --- | --- | --- | --- |
| soft label 1 ($\mathcal{Y}'_1$) | **0.3645** | **0.3645** | 0.2252 | 0.2252 | **0.6172** |
| soft label 2 ($\mathcal{Y}'_2$) | **0.3645** | **0.3645** | 0.2252 | 0.2252 | **0.6172** |
| soft label 3 ($\mathcal{Y}'_3$) | 0.1900 | 0.1900 | **0.2456** | **0.2456** | 0.4350 |
| soft label 4 ($\mathcal{Y}'_4$) | 0.1900 | 0.1900 | **0.2456** | **0.2456** | 0.4350 |
| soft label 5 ($\mathcal{Y}'_5$) | 0.1252 | 0.1252 | 0.1260 | 0.1260 | **0.4480** |

N.B. $\mathcal{Y}_1 = \mathcal{Y}_2$ and $\mathcal{Y}_3 = \mathcal{Y}_4$. $\mathcal{Y}_5 = (\neg \mathcal{Y}_1) \wedge (\neg \mathcal{Y}_2) \wedge (\neg \mathcal{Y}_3) \wedge (\neg \mathcal{Y}_4)$.

TABLE VI
INFLUENCE WEIGHTS ($S$) OF ORIGINAL LABELS ON A SOFT LABEL IN UNION DATASET

|  | original label 1 ($\mathcal{Y}_1$) | original label 2 ($\mathcal{Y}_2$) | original label 3 ($\mathcal{Y}_3$) | original label 4 ($\mathcal{Y}_4$) | original label 5 ($\mathcal{Y}_5$) |
| --- | --- | --- | --- | --- | --- |
| soft label 1 ($\mathcal{Y}'_1$) | **0.2295** | 0.0798 | 0.0981 | 0.1206 | **0.2654** |
| soft label 2 ($\mathcal{Y}'_2$) | 0.0791 | **0.1529** | 0.0363 | 0.0551 | **0.1327** |
| soft label 3 ($\mathcal{Y}'_3$) | 0.1378 | 0.0520 | **0.1694** | 0.1017 | **0.2151** |
| soft label 4 ($\mathcal{Y}'_4$) | 0.0077 | -0.0002 | 0.0005 | **0.0668** | 0.0106 |
| soft label 5 ($\mathcal{Y}'_5$) | 0.0649 | -0.0107 | -0.0264 | 0.0351 | **0.1057** |

N.B. $\mathcal{Y}_1 = \mathcal{Y}_2 \vee \mathcal{Y}_3 \vee \mathcal{Y}_4$ and $\mathcal{Y}_5 = (\neg \mathcal{Y}_1) \wedge (\neg \mathcal{Y}_2) \wedge (\neg \mathcal{Y}_3) \wedge (\neg \mathcal{Y}_4)$.

Tables IV-VI, we can find that the influence of $\mathcal{Y}_5$ on the soft label $\mathcal{Y}'_5$ is most significant, whereas the influence of $\mathcal{Y}_1 \sim \mathcal{Y}_4$ on $\mathcal{Y}'_5$ is relatively small.

2) In Table IV, the first four labels $\mathcal{Y}_1$, $\mathcal{Y}_2$, $\mathcal{Y}_3$ and $\mathcal{Y}_4$ are independent of each other, and $\mathcal{Y}_5 = (\neg \mathcal{Y}_1) \wedge (\neg \mathcal{Y}_2) \wedge (\neg \mathcal{Y}_3) \wedge (\neg \mathcal{Y}_4)$. The results of influence weights in Table IV show that the effect of $\mathcal{Y}_1$, $\mathcal{Y}_2$, $\mathcal{Y}_3$ and $\mathcal{Y}_4$ on the soft labels $\mathcal{Y}'_1$, $\mathcal{Y}'_2$, $\mathcal{Y}'_3$ and $\mathcal{Y}'_4$, respectively, are significant. In addition, the contribution of $\mathcal{Y}_5$ to $\mathcal{Y}'_1$, $\mathcal{Y}'_2$, $\mathcal{Y}'_3$ and $\mathcal{Y}'_4$ is also obvious.

3) In Table V, $\mathcal{Y}_1 = \mathcal{Y}_2$, $\mathcal{Y}_3 = \mathcal{Y}_4$, and $\mathcal{Y}_5 = (\neg \mathcal{Y}_1) \wedge (\neg \mathcal{Y}_2) \wedge (\neg \mathcal{Y}_3) \wedge (\neg \mathcal{Y}_4)$. The results of influence weights in Table V reveal that $\mathcal{Y}_5$ has a greater influence on the soft labels $\mathcal{Y}'_1$, $\mathcal{Y}'_2$, $\mathcal{Y}'_3$ and $\mathcal{Y}'_4$. Meanwhile, it is obvious that $\mathcal{Y}_1$ and $\mathcal{Y}_2$ have the same influence on $\mathcal{Y}'_1$ ($\mathcal{Y}'_2$), and $\mathcal{Y}_3$ and $\mathcal{Y}_4$ have the same influence on $\mathcal{Y}'_3$ ($\mathcal{Y}'_4$).

4) In Table VI, $\mathcal{Y}_1 = \mathcal{Y}_2 \vee \mathcal{Y}_3 \vee \mathcal{Y}_4$ and $\mathcal{Y}_5 = (\neg \mathcal{Y}_1) \wedge (\neg \mathcal{Y}_2) \wedge (\neg \mathcal{Y}_3) \wedge (\neg \mathcal{Y}_4)$. From the results of influence weights in Table VI, we can see that it is $\mathcal{Y}_1$ and $\mathcal{Y}_5$ that affect the soft label $\mathcal{Y}'_1$ significantly, and that the effect of $\mathcal{Y}_2 \sim \mathcal{Y}_4$ on the soft label $\mathcal{Y}'_1$ are similar.

The above findings are consistent with the logical relationship we designed for the labels, which validates that the soft label learning in R-MLTSK-FS is effective.

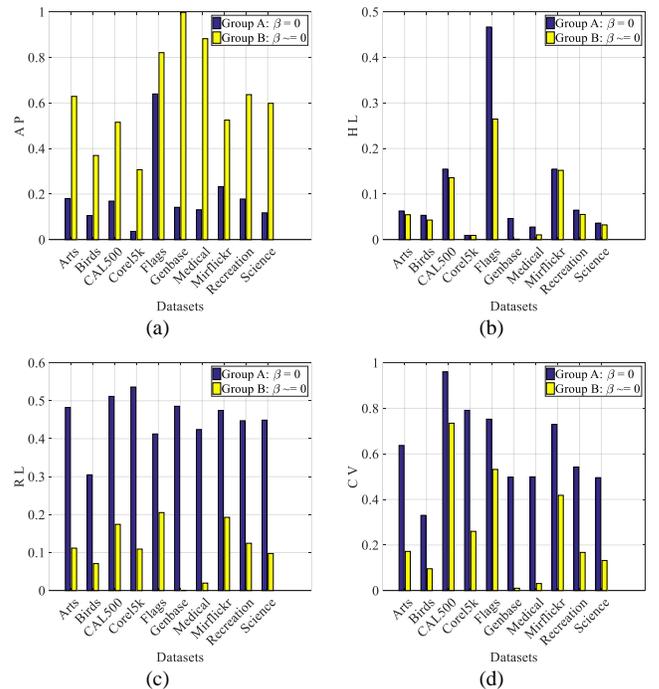

Fig. 3 Effect of soft label learning on the performance of the proposed R-MLTSK-FS. (Soft label learning is disabled for Group A and enabled in

Group B. For AP, the larger the value, the better the performance; for HL, RL and CV, the smaller the value, the better the performance.)

*2) Ablation Experiment Analysis for Soft Label Learning:* The influence of soft label learning on the performance of the proposed method is analyzed by setting $\beta$ in Eq. (12) to 0. Specifically, we conduct two groups of experiments, Group A and Group B, on 10 datasets. In Group A, soft label learning is disabled. That is, $\beta$ is set to 0. In Group B, soft label learning is enabled and $\beta$ is an appropriate value. For other parameters, their settings in Group A and Group B are the same. The experiment results in terms of AP, HL, RL and CV are shown in Fig. 3. We can find that the performance in Group A is significantly lower than that in Group B, which indicates that soft label learning has a critical impact on the learning of the proposed method.

*(4) Effectiveness Analysis of Correlation Enhancement Learning*

In this section, we conduct visualization analysis and ablation analysis for the proposed correlation enhancement learning to verify the effectiveness of the mechanism. The details are as follows:

*1) Visualization Analysis for Correlation Enhancement Learning:* In order to verify the effectiveness of the correlation enhancement learning mechanism in guiding the consequent vector optimization, we conduct correlation visualization experiment on the Science dataset, where the dimension of label space is 40. Specifically, the Pearson correlation coefficient is used to measure the correlation between two vectors. The higher the value of Pearson correlation coefficient, the stronger the correlation between two vectors. Experimental results are shown in Fig. 4, where Fig. 4(a) visualizes the correlation between any two original labels, and Fig. 4(b) visualizes the correlation between any two optimized consequent vectors associated with the corresponding labels. For an effective correlation enhancement learning mechanism, the correlation coefficient between two consequent vectors should be kept close to that between their corresponding labels.

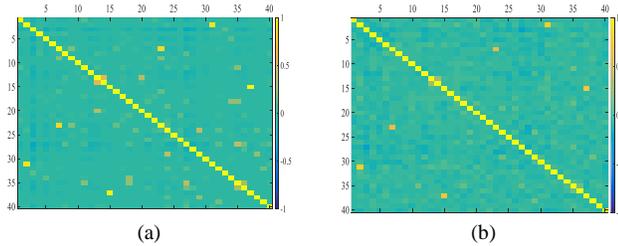

Fig. 4 Visualization of label correlation learning on the Science dataset: (a) visualization of the correlation coefficient between any two original label vectors, and (b) visualization of the correlation coefficient between any two consequent vectors associated with the corresponding labels. The higher the value of correlation coefficient, the stronger the correlation between two vectors.

It is clear that there is little difference between Fig. 4(a) and Fig. 4(b), indicating that the correlation between the labels can closely guide the learning of the corresponding consequent vectors, and demonstrating the effectiveness of the correlation enhancement learning mechanism.

*2) Ablation Analysis for Correlation Enhancement Learning:* Two groups of experiments, Group A and Group B, are performed on the 10 datasets to analyze the effect of the correlation enhancement learning. The trade-off parameter $\gamma$ in Eq. (12) is set to 0 in Group A to disable the correlation enhancement learning mechanism. The mechanism is enabled in Group B with $\gamma$ set to an appropriate value. The setting of the other trade-off parameters in two groups are the same. Experimental results in terms of AP, HL, RL and CV are shown in Fig. 5. We can find that compared with Group A, the performance in Group B is better, which indicates that the proposed correlation enhancement learning can improve the performance of the proposed R-MLTSK-FS.

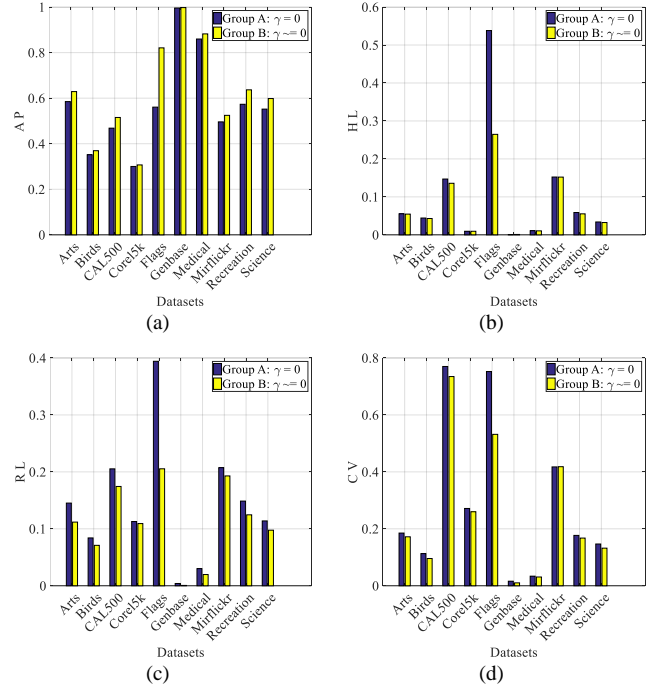

Fig. 5 Effect of correlation enhancement learning on the performance of the proposed R-MLTSK-FS. (Correlation enhancement learning is disabled in Group A and enabled in Group B. For AP, the larger the value, the better the performance; for HL, RL and CV, the smaller the value, the better the performance.)

*(5) Cross-Validation based Parameter Analysis*

In this section, we analyze the influence of the hyperparameters $\alpha$, $\beta$, $\gamma$ and $K$ on the classification performance of R-MLTSK-FS in terms of AP. In the analysis, we study the sensitivity of the classification performance to one of the four hyperparameter by keeping the other three fixed. For example, we fix the values of $\beta$, $\gamma$ and $K$, and adjust the value of $\alpha$ to analyze the effect of $\alpha$. The hyperparameters $\alpha$, $\beta$ and $\gamma$ are varied within $\{10^{-3}, 10^{-2}, 10^{-1}, 10^{0}, 10^{1}, 10^{2}\}$ and $K$ is varied within $\{2, 3, 4, 5, 6, 7, 8, 9, 10\}$. The AP values of R-MLTSK-FS are obtained with five-fold cross-validation strategy. The experimental results are shown in Fig. 6, from which the following observations are obtained:

1) When $\alpha$ is in the range of $(10^{-3}, 10^{0})$, the performance of R-MLTSK-FS in terms of AP is stable for most datasets. In addition, AP decreases with increasing $\alpha$ for most datasets when $\alpha$ is within $(10^{1}, 10^{2})$. For the CAL500 dataset, AP increases with $\alpha$. In general, R-MLTSK-FS is stable and can achieve optimal performance when $\alpha$ is in the range of $(10^{-2}, 10^{0})$.

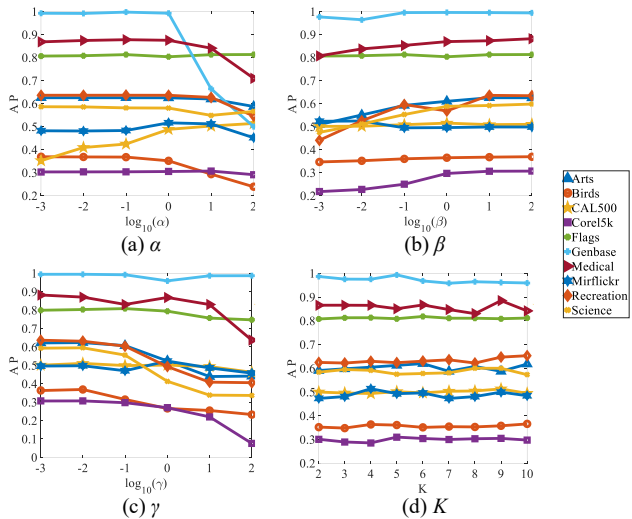

Fig. 6 The influence of the hyperparameters (a) $\alpha$, (b) $\beta$, (c) $\gamma$, and (d) $K$ on AP of the R-MLTSK-FS.

2) In general, R-MLTSK-FS is sensitive to $\beta$ when it is in the range of $(10^{-3}, 10^0)$. It is stable and can reach an optimal AP value for the 10 datasets when $\beta$ is within $(10^1, 10^2)$.

3) For the hyperparameter $\gamma$, AP fluctuates in a similar way for all the 10 datasets. In general, the performance of R-MLTSK-FS is stable when $\gamma$ is within $(10^{-3}, 10^{-1})$. The AP value fluctuates significantly when $\gamma$ is in the range of $(10^{-1}, 10^2)$, while exhibiting a decreasing trend with increasing $\gamma$. In general, optimal AP can be achieved for all the 10 datasets when $\gamma$ is in the range of $(10^{-3}, 10^{-1})$.

4) The AP value for the 10 datasets fluctuates slightly with increasing $K$. Optimal values of AP can be obtained when $K$ is within $(4, 9)$.

According to the above analysis, the appropriate values of $\alpha$, $\beta$, $\gamma$ and $K$ can be searched in the ranges of $(10^{-2}, 10^0)$, $(10^1, 10^2)$, $(10^{-3}, 10^{-1})$ and $(4, 9)$ respectively, to obtain a stable and outstanding performance for the proposed R-MLTSK-FS.

*(6) Convergence Analysis*

The Birds and Flags datasets are adopted in this part to investigate the convergence of the proposed method. The results are shown in Fig. 7, where the vertical axis is the absolute value of the difference between the previous and the current value of the objective function (denoted by df), and the horizontal axis is the number of iterations. It can be seen from Fig. 7 that R-MLTSK-FS can converge within 10 iterations for the two datasets.

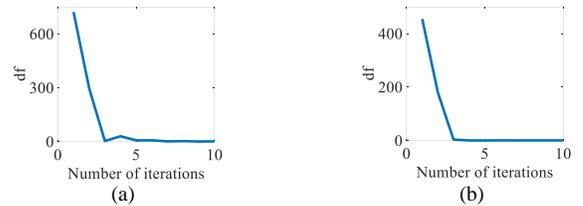

Fig. 7 Convergence analysis for datasets (a) Birds and (b) Flags.

*(7) Statistical Analysis*

We employ the Friedman test and the Bonferroni-Dunn test to evaluate the statistical significance of the difference observed between the proposed R-MLTSK-FS and the eleven comparison methods [56]. The details are as follows.

*1) Friedman Test:* Based on the performance evaluation results in Table III, we perform the Friedman test in terms of AP, HL, RL and CV. The null hypothesis is that there is no significant difference between all the methods in terms of the four metrics. For each metric, if the Friedman statistic $F_F$ is greater than a critical value (i.e., 1.8867), the null hypothesis for that metric is rejected, which means the difference is statistically significant. The results of the Friedman test, corresponding to the results in Table III, are shown in Table VII. It can be seen from Table VII that the null hypotheses on AP, HL, RL and CV are all rejected. This means that the differences in classification performance of all the twelve methods (including our R-MLTSK-FS) are significant in terms of the four metrics. Next, we conduct the post-hoc Bonferroni-Dunn test to evaluate whether the difference in performance between R-MLTSK-FS and the comparison methods is statistically significant.

TABLE VII
FRIEDMAN STATISTICS

| Evaluation metric | $F_F$ | Critical value ($\alpha = 0.05$) |
|---|---|---|
| AP | 21.1229 | |
| HL | 4.6842 | 1.8867 |
| RL | 13.9637 | |
| CV | 28.5657 | |

TABLE VIII
MEAN PERFORMANCE (RANKING) OF MULTILABEL CLASSIFICATION METHODS IN TERMS OF AP

|  | BR | MLkNN | MLSF | ML-TSK FS | CC | RAkEL | CorrLog | HNOML | ML-LSS | 2SML | RLFSCL | R-MLTSK-FS |
|---|---|---|---|---|---|---|---|---|---|---|---|---|
| Arts | 0.6270 (3) | 0.5454 (9) | 0.4977 (10) | 0.6207 (5) | 0.6164 (6) | 0.2682 (12) | 0.3646 (11) | 0.6090 (7) | 0.6273 (2) | 0.6247 (4) | 0.5810 (8) | 0.6289 (1) |
| Birds | 0.3422 (4) | 0.2303 (11) | 0.2712 (10) | 0.3438 (3) | 0.3360 (5) | 0.3591 (2) | 0.2124 (12) | 0.3352 (6) | 0.3230 (7) | 0.2769 (9) | 0.2965 (8) | 0.3694 (1) |
| CAL500 | 0.5048 (5) | 0.4965 (6) | 0.4906 (7) | 0.5075 (3) | 0.4541 (8) | 0.2150 (12) | 0.3108 (11) | 0.4314 (10) | 0.5095 (2) | 0.4460 (9) | 0.5069 (4) | 0.5153 (1) |
| Corel5k | 0.3044 (5) | 0.2561 (9) | 0.2134 (10) | 0.3064 (2) | 0.2639 (8) | 0.0652 (12) | 0.2079 (11) | 0.2884 (6) | 0.3051 (4) | 0.2843 (7) | 0.3058 (3) | 0.3070 (1) |
| Flags | 0.8101 (6) | 0.8020 (9) | 0.8163 (4) | 0.8176 (3) | 0.8076 (8) | 0.6581 (12) | 0.7704 (10) | 0.8080 (7) | 0.8182 (2) | 0.6921 (11) | 0.8148 (5) | 0.8209 (1) |
| Genbase | 0.9922 (6) | 0.9910 (8) | 0.9913 (7) | 0.9968 (2) | 0.9802 (10) | 0.7784 (12) | 0.9717 (11) | 0.9941 (5) | 0.9967 (3) | 0.9955 (4) | 0.9886 (9) | 0.9977 (1) |
| Medical | 0.8755 (7) | 0.8067 (10) | 0.8272 (9) | 0.8959 (2) | 0.8765 (5) | 0.4443 (12) | 0.7562 (11) | 0.8761 (6) | 0.8822 (3.5) | 0.9059 (1) | 0.8386 (8) | 0.8822 (3.5) |
| Mirflickr | 0.4540 (10) | 0.5096 (6) | 0.2906 (11) | 0.5239 (2) | 0.4703 (9) | 0.2216 (12) | 0.4779 (8) | 0.5121 (5) | 0.5222 (3) | 0.5201 (4) | 0.5053 (7) | 0.5246 (1) |
| Recreation | 0.6363 (2) | 0.5333 (9) | 0.4817 (10) | 0.6362 (3) | 0.6286 (6) | 0.2922 (11) | 0.2104 (12) | 0.6062 (7) | 0.6287 (5) | 0.6297 (4) | 0.5901 (8) | 0.6366 (1) |
| Science | 0.5983 (2) | 0.5134 (9) | 0.4461 (10) | 0.5978 (3) | 0.5861 (6) | 0.2333 (12) | 0.2492 (11) | 0.5737 (7) | 0.5918 (5) | 0.5963 (4) | 0.5567 (8) | 0.5984 (1) |
| *Aver. Ranking* | 5 | 8.6 | 8.8 | 2.8 | 7.1 | 10.9 | 10.8 | 6.6 | 3.65 | 5.7 | 6.8 | 1.25 |

TABLE IX
AVERAGE RANKINGS OF MULTILABEL CLASSIFICATION METHODS IN TERMS OF HL, RL AND CV

|    | BR  | ML$k$NN | MLSF | ML-TSK FS | CC   | RA$k$EL | CorrLog | HNOML | ML-LSS | 2SML | RLF-SCL | R-MLTSK-FS |
|----|-----|---------|------|-----------|------|---------|---------|-------|--------|------|---------|------------|
| HL | 8.2 | 7.85    | 7.65 | 3.8       | 8.95 | 8.3     | 6.8     | 6.95  | 4.45   | 6.95 | 6.25    | 1.85       |
| RL | 6.7 | 7.3     | 5.5  | 3.9       | 7.7  | 11.5    | 10.4    | 5.35  | 3.9    | 6.4  | 8       | 1.35       |
| CV | 6.1 | 6.2     | 8.1  | 2.5       | 7.8  | 11.2    | 8.9     | 5.8   | 3      | 6.5  | 10.8    | 1.1        |

*2) Bonferroni-Dunn Test:* According to the results in Friedman test, we conduct the post-hoc test based on the results of AP, HL, RL and CV respectively, where R-MLTSK-FS is set as the control method.

Firstly, we calculate the average ranking of the 12 methods for each metric respectively. According to the AP metric in Table III, the average performance ranking of the twelve methods are shown in Table VIII. The number inside bracket "()" is the ranking (i.e., sorting order) of a method for the same dataset. The last row in Table VIII shows the average ranking of each method in terms of AP performance. The smaller the average ranking, the better the method in terms of AP. Similarly, the average ranking of 12 methods on HL, RL and CV are obtained and shown in Table IX.

Secondly, we calculate the critical difference (CD), which is a standard used for evaluating the difference in average ranking between the methods, using the equation below:

$$\text{CD} = q_\alpha \sqrt{n(n+1)/6M} \tag{38}$$

where $n$ and $M$ are the number of methods ($n = 12$) and the number of datasets ($M = 10$), respectively. With confidence level $\alpha = 0.05$ and $q_\alpha = 3.268$, we have CD = 5.2695.

Fig. 8 gives the average rankings of the twelve methods, which are shown on the horizontal line with ticks marking 1 to 12. The smaller the average ranking (i.e., closer to the right), the better the method. As R-MLTSK-FS is at the rightmost position on the horizontal line, for all the four metrics, it is the best among the twelve methods. A red line of length one CD is drawn from R-MLTSK-FS to the left. For a method located within the span of the red line, the difference in average ranking between the method and R-MLTSK-FS is less than one CD, indicating that the performance difference between them is small. Otherwise, the difference is significant. The following conclusions can be drawn from Fig. 8. Firstly, the performance of CC and RA$k$EL are significantly lower than that of R-MLTSK-FS in terms of the four metrics. Secondly, for BR, CorrLog, HNOML, ML$k$NN, MLSF and RLFSCL, their performance is mediocre. Thirdly, for ML-LSS, ML-TSK FS and 2SML, their performance is not significantly lower than that of R-MLTSK-FS. Fourthly, R-MLTSK-FS is superior to other methods on the four metrics.

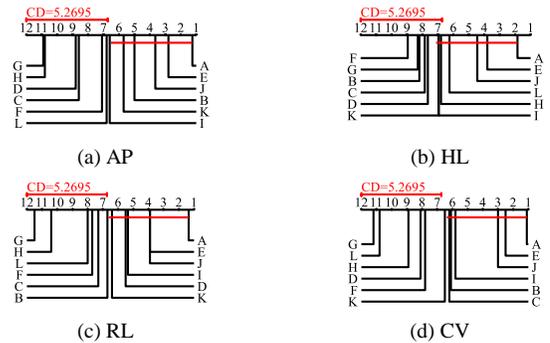

Fig. 8 Comparison of R-MLTSK-FS (as control) with the other methods using the Bonferroni-Dunn test. (The letter A refers to R-MLTSK-FS, B to BR, C to ML$k$NN, D to MLSF, E to ML-TSK FS, F to CC, G to RA$k$EL, H to CorrLog, I to HNOML, J to ML-LSS, K to 2SML, and L to RLFSCL respectively.)

*(8) Running Time Analysis*

Further to the discussion of the computational complexity of the 12 methods in Section IV-C, experiments are conducted to measure and analyze their running time. In the experiments, grid search is not considered and parameter optimization is only performed once for each dataset. Five-fold cross strategy is still adopted to obtain the stable results. The results in Table X show that the proposed R-MLTSK-FS has no advantage in running speed. This is mainly due to the fact that R-MLTSK-FS takes into account the impact of all original labels on a soft label, it is necessary to learn from the original label space for $L$ times in order to construct the soft label space. Besides, since R-MLTSK-FS maps all the original features into the fuzzy feature space, some redundant or unimportant features may exist and slow down the process. Further research will be conducted to improve the speed of the proposed method while maintaining performance.

*(9) Interpretability Analysis*

Interpretable methods can be divided into two categories, i.e., post-hoc methods (such as black box-based deep networks) and intrinsically interpretable methods (such as rule-based methods) [67]. The former interprets a model by analyzing or adjusting the parameters, whereas the latter interprets a model through a transparent inference process. Since the proposed R-MLTSK-FS is based on fuzzy rules to achieve transparency, it to the category of intrinsically interpretable methods. At present, a consensus on the suitable metrics to evaluate the transparency is not available yet. However, we can refer to the interpretability analysis in [68] to demonstrate and evaluate the interpretability of the proposed method.

Based on the above analysis, we take the Flags dataset as an example to show the transparency and interpretability of the proposed R-MLTSK-FS. The classification task of the Flags dataset is to predict the religion of a country by analyzing the area, population and flag information. Specifically, we set $K = 3$, $\alpha = 0.1$, $\beta = 10$ and $\gamma = 0.001$. For illustration purpose and simplicity, we only show the linguistic inference of the 3rd, 4th and 8th features on the 1st and 7th labels. Fuzzy parameters obtained by R-MLTSK-FS on the Flags dataset are shown in Table XI, where the fuzzy antecedent parameters and fuzzy consequent parameters under different rules are included. Based on the obtained fuzzy parameters in Table XI, Fig. 9 visualizes the fuzzy sets of the 3rd, 4th and 8th features under the three rules.

For each feature, its antecedent fuzzy sets (i.e., antecedent parameters) obtained under different rules have different linguistic interpretations. For example, the fuzzy set of "Feature 3 (Area)" corresponds to "Small" under the 1st rule, "Medium" under the 2nd rule and "Large" under the 3rd rule. Therefore, combined with the feature semantics under the different rules, the inference outputs of R-MLTSK-FS under different rules are expressed as follows:

TABLE X
MEAN (SD) OF THE RUNNING TIME (SECONDS) OF MULTILABEL CLASSIFICATION METHODS

|  | BR | MLkNN | MLSF | ML-TSK FS | CC | RAkEL | CorrLog | HNOML | ML-LSS | 2SML | RLFSCL | R-MLTSK-FS |
|---|---|---|---|---|---|---|---|---|---|---|---|---|
| Arts | 47.0681 (0.0755) | 24.9206 (0.4397) | 331.2876 (10.2956) | 0.8306 (0.0243) | 49.4015 (0.0466) | 0.7620 (0.1825) | 5.0194 (0.0322) | 109.5474 (0.2911) | 13.9256 (3.5510) | 4.6331 (0.0761) | 1.8315 (0.0679) | 68.7440 (0.2089) |
| Birds | 0.1959 (0.0343) | 0.3672 (0.0447) | 3.0118 (0.1319) | 0.0411 (0.0093) | 0.1907 (0.0324) | 0.1309 (0.0318) | 0.5001 (0.0174) | 2.0969 (0.2554) | 0.4109 (0.0191) | 0.2284 (0.0109) | 0.0215 (0.0023) | 6.6382 (0.2222) |
| CAL500 | 0.0344 (0.0069) | 0.4900 (0.0598) | 17.8397 (0.1698) | 0.1384 (0.0191) | 0.1000 (0.0057) | 0.3609 (0.0080) | 2.9862 (0.0637) | 2.0130 (0.4384) | 2.0487 (0.0131) | 0.4602 (0.0457) | 0.1734 (0.0024) | 1.5332 (1.5182) |
| Corel5k | 295.0713 (5.4033) | 31.2489 (0.2326) | 4241.7109 (290.8981) | 8.8909 (0.2368) | 1198.6807 (92.9728) | 3.1677 (0.4124) | 132.4079 (2.0447) | 171.2488 (0.2179) | 219.4436 (0.6736) | 10.5249 (0.0980) | 15.4731 (0.0653) | 80.0878 (50.0149) |
| Flags | 0.0071 (0.0062) | 0.1834 (0.1415) | 0.2024 (0.0983) | 0.0171 (0.0105) | 0.0037 (0.0005) | 0.0017 (0.0006) | 0.1269 (0.0924) | 0.1057 (0.1525) | 0.0119 (0.0005) | 0.0786 (0.0199) | 0.0052 (0.0035) | 0.0874 (0.0229) |
| Genbase | 72.3212 (1.7812) | 0.9731 (0.1200) | 4.5561 (0.1291) | 3.5131 (0.2509) | 76.0524 (1.8687) | 0.0060 (0.0005) | 1.1140 (0.0244) | 6.9896 (4.4126) | 40.7077 (2.6764) | 2.2976 (0.0308) | 1.2913 (0.0274) | 172.3668 (79.5331) |
| Medical | 161.4918 (8.7422) | 3.3133 (0.1412) | 17.2766 (0.4413) | 2.9983 (0.2047) | 184.6018 (8.3833) | 0.0380 (0.0501) | 2.7784 (0.1271) | 16.8962 (17.0204) | 612.3094 (18.9868) | 14.7315 (0.2076) | 2.6951 (0.0462) | 434.8960 (153.6957) |
| Mirflickr | 429.7999 (16.4332) | 375.8408 (3.4309) | 6828.5391 (85.8694) | 1.1883 (0.0160) | 382.7079 (12.6850) | 692.5505 (28.4520) | 19.1700 (0.4320) | 971.9104 (167.0872) | 485.4434 (52.1929) | 33.9120 (0.1424) | 11.7832 (0.0668) | 519.8169 (3.6779) |
| Recreation | 66.8139 (3.6614) | 31.6512 (0.5623) | 279.7403 (6.5017) | 1.1726 (0.0434) | 68.2004 (0.9738) | 5.7419 (0.1334) | 4.9813 (0.0855) | 146.8607 (0.3426) | 37.7360 (2.8824) | 5.7051 (0.0637) | 2.5788 (0.0785) | 104.8956 (0.1690) |
| Science | 85.9245 (0.6993) | 39.5522 (1.1009) | 486.3910 (8.6843) | 2.2019 (0.0563) | 93.1481 (0.8251) | 6.0021 (0.0031) | 9.1343 (0.1691) | 181.4162 (3.4490) | 65.0359 (1.9975) | 7.4437 (0.0691) | 4.2783 (0.0658) | 154.6031 (0.9552) |

TABLE XI
FUZZY SYSTEM PARAMETERS WITH THREE RULES OF R-MLTSK-FS ON THE FLAGS DATASET

| Rule Base |
|---|
| IF: $x_1$ is $A_1^k$ $(m_1^k, \delta_1^k) \wedge x_2$ is $A_2^k$ $(m_2^k, \delta_2^k) \wedge \cdots \wedge x_D$ is $A_D^k$ $(m_D^k, \delta_D^k)$, |
| THEN: $f^k(x) = \begin{bmatrix} f_1^k(x) \\ \vdots \\ f_l^k(x) \\ \vdots \\ f_L^k(x) \end{bmatrix} = \begin{bmatrix} c_{10}^k + c_{11}^k x_1 + \cdots + c_{1D}^k x_D \\ \vdots \\ c_{l0}^k + c_{l1}^k x_1 + \cdots + c_{lD}^k x_D \\ \vdots \\ c_{L0}^k + c_{L1}^k x_1 + \cdots + c_{LD}^k x_D \end{bmatrix}$ |

| Rule No. | Antecedent Parameter | Consequent Parameter |
|---|---|---|
| $k$ | $\boldsymbol{m}^k = [m_1^k, m_2^k, \ldots, m_D^k]$ <br> $\boldsymbol{\delta}^k = [\delta_1^k, \delta_2^k, \ldots, \delta_D^k]$ | $\boldsymbol{c}_l^k = [c_{l0}^k, c_{l1}^k, \ldots, c_{lD}^k]^{\mathrm{T}}$ |
| 1 | $\boldsymbol{m}^1 = [\ldots, 116.7219, 8.3576, \ldots, 1.5430, \ldots]$ <br> $\boldsymbol{\delta}^1 = [\ldots, 1.0000, 1.9620, \ldots, 3.8456, \ldots]$ | $\boldsymbol{c}_1^1 =$ <br> $[0.9784, \ldots, -0.0115, -0.0049, \ldots, 0.0306, \ldots]^{\mathrm{T}}$ <br> $\vdots$ <br> $\boldsymbol{c}_7^1 =$ <br> $[-0.1723, \ldots, 0.0004, 0.0015, \ldots 0.0043, \ldots]^{\mathrm{T}}$ |
| 2 | $\boldsymbol{m}^2 = [\ldots, 1.1551e + 03, 25.0313, \ldots, 1.3438, \ldots]$ <br> $\boldsymbol{\delta}^2 = [\ldots, 1.0723, 1.9134, \ldots, 3.8732, \ldots]$ | $\boldsymbol{c}_1^2 =$ <br> $[0.0000, \ldots, 0.0005, 0.0000, \ldots, 0.0000, \ldots]^{\mathrm{T}}$ <br> $\vdots$ <br> $\boldsymbol{c}_7^2 =$ <br> $[0.0000, \ldots, 0.0000, 0.0000, \ldots, 0.0000, \ldots]^{\mathrm{T}}$ |
| 3 | $\boldsymbol{m}^3 = [\ldots, 5.8818e + 03, 217.7000, \ldots, 2.5000, \ldots]$ <br> $\boldsymbol{\delta}^3 = [\ldots, 10.0000, 8.1969, \ldots, 4.3534, \ldots]$ | $\boldsymbol{c}_1^3 =$ <br> $[0.0000, \ldots, 0.0000, 0.0000, \ldots, 0.0000, \ldots]^{\mathrm{T}}$ <br> $\vdots$ <br> $\boldsymbol{c}_7^3 =$ <br> $[0.0000, \ldots, 0.0000, 0.0000, \ldots, 0.0000, \ldots]^{\mathrm{T}}$ |

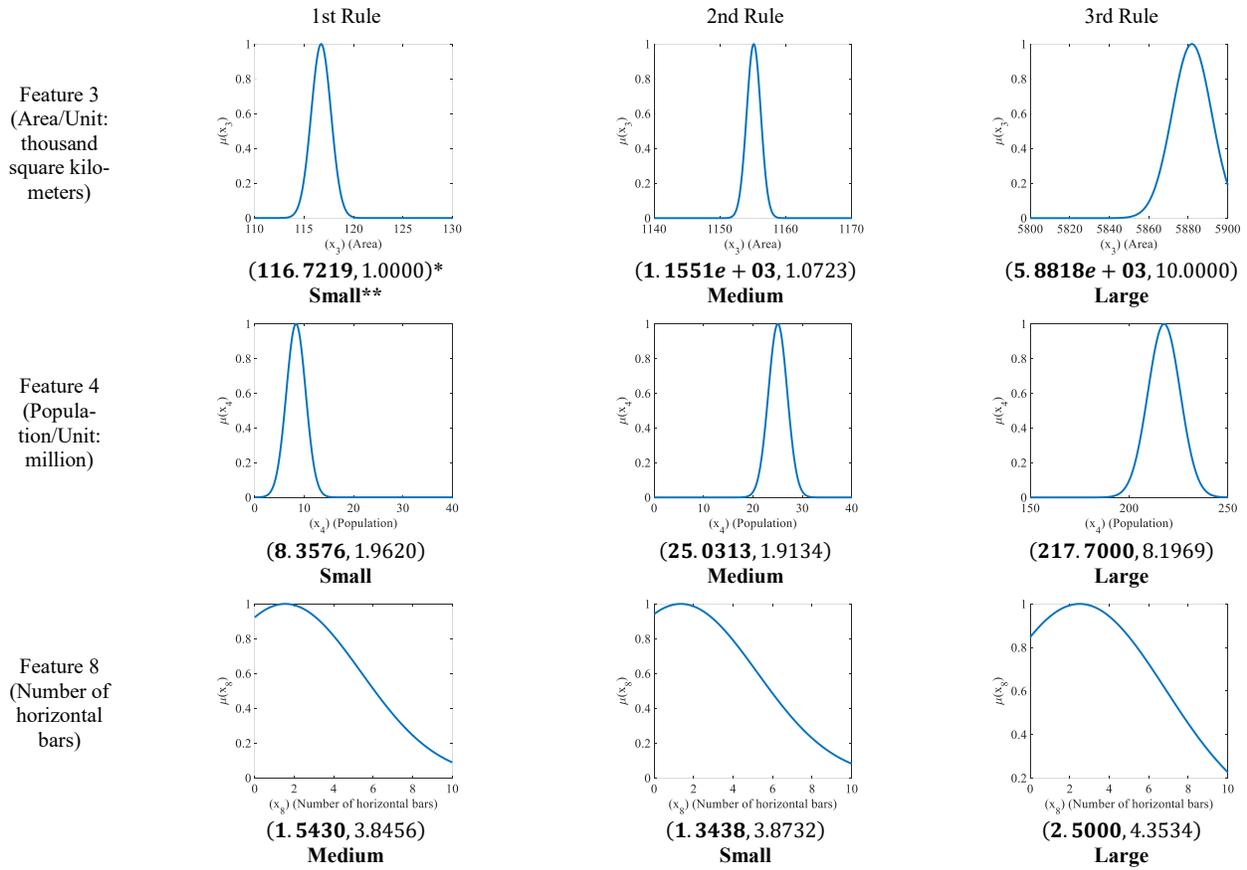

Fig. 9 Antecedent fuzzy sets of the 3rd, 4th and 8th features in the Flags dataset and their linguistic interpretations when the number of rules is three. (* denotes the antecedent parameter of feature 3 under the first rule, and ** denotes the explanation for the corresponding fuzzy set).

*The first rule:*
IF feature1 is ...,
AND ...
AND feature3 (Area) is Small,
AND feature4 (Population) is Small,
AND ...
AND feature8 (Number of horizontal bars) is Medium,
AND ...
THEN this rule gives the output below:

$$f^1(x) = \begin{bmatrix} f_1^1(x) \\ \vdots \\ f_7^1(x) \end{bmatrix} =$$

$$\begin{bmatrix} 0.9784 + \cdots - 0.0115x_3 - 0.0049x_4 + \cdots + 0.0306x_8 + \cdots \\ \vdots \\ -0.1723 + \cdots + 0.0004x_3 + 0.0015x_4 + \cdots + 0.0043x_8 + \cdots \end{bmatrix}$$

*The second rule:*
IF feature1 is ...,
AND ...
AND feature3 (Area) is Medium,
AND feature4 (Population) is Medium,
AND ...
AND feature8 (Number of horizontal bars) is Small,
AND ...
THEN this rule gives the output below:

$$f^2(x) = \begin{bmatrix} f_1^2(x) \\ \vdots \\ f_7^2(x) \end{bmatrix} =$$

$$\begin{bmatrix} 0.0000 + \cdots + 0.0005x_3 + 0.0000x_4 + \cdots + 0.0000x_8 + \cdots \\ \vdots \\ 0.0000 + \cdots + 0.0000x_3 + 0.0000x_4 + \cdots + 0.0000x_8 + \cdots \end{bmatrix}$$

*The third rule:*
IF feature1 is ...,
AND ...
AND feature3 (Area) is Large,
AND feature4 (Population) is Large,
AND ...
AND feature8 (Number of horizontal bars) is Large,
AND ...
THEN this rule gives the output below:

$$f^3(x) = \begin{bmatrix} f_1^3(x) \\ \vdots \\ f_7^3(x) \end{bmatrix} =$$

$$\begin{bmatrix} 0.0000 + \cdots + 0.0000x_3 + 0.0000x_4 + \cdots + 0.0000x_8 + \cdots \\ \vdots \\ 0.0000 + \cdots + 0.0000x_3 + 0.0000x_4 + \cdots + 0.0000x_8 + \cdots \end{bmatrix}$$

The above process shows the transparency of the proposed R-MLTSK-FS which is attributable to the rule-based fuzzy inference. That is, R-MLTSK-FS first utilizes the IF-parts of fuzzy rules to divide the feature data into different antecedent fuzzy sets (i.e., infusing linguistic interpretations of different levels), and then utilizes the THEN-parts of fuzzy rules to model the inference relationship between features and labels. The IF-THEN rules correspond to different linguistic interpretations for the inference process. On the whole, the transparency of the R-MLTSK-FS is demonstrated experimentally and the proposed method is semantically interpretable to a certain extent.

## V. CONCLUSION

The robust multilabel learning method R-MLTSK-FS with strong fuzzy inference ability, label correlation learning ability and robustness against noisy labels is proposed in this paper. From the aspect of soft label learning, R-MLTSK-FS constructs the soft label space to reduce the influence of label noise. From the aspect of soft multilabel loss function construction, R-MLTSK-FS utilizes the fuzzy rule-based multi-output TSK FS as a transparent model to build the inference relationship between input features and soft labels, and then the loss function is constructed based on the multi-output TSK FS and soft labels to enhance model training. From the aspect of correlation enhancement learning, R-MLTSK-FS utilizes the correlation information between soft labels to constrain the learning of model parameters and enhance the learning ability. Experimental analyses on ten benchmark multilabel datasets and three synthetic multilabel datasets show the promising performance of R-MLTSK-FS.

Further research on R-MLTSK-FS will proceed along two directions. First, we will reduce the complexity of soft label learning. Since R-MLTSK-FS considers all the original labels for a soft label, which is computationally intensive, research will be conducted to model with random label subsets for a soft label to reduce the complexity. Second, we will simplify the rule base of TSK FS. In R-MLTSK-FS, the fuzzy system transforms all the original features into the fuzzy feature space. If the dimension of the original feature space is large, the learning speed of R-MLTSK-FS will be slow. Hence, a screening mechanism such as feature reduction in [90, 91] will be developed to identify representative subsets of the original features to improve the learning efficiency.

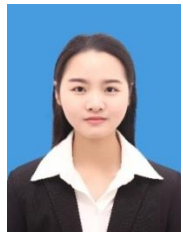

**Qiongdan Lou** received the B.S. degree in software engineering from Jiangsu University of Science and Technology, Suzhou, China, in 2017, and the Ph.D. degree in the School of Artificial Intelligence and Computer Science, Jiangnan University, Wuxi, China, in 2023.

Her research interests include interpretability and uncertainty artificial intelligence and pattern recognition.

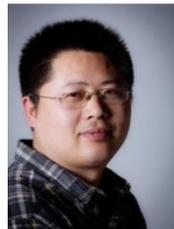

**Zhaohong Deng** (M'12-SM'14) received the B.S. degree in physics from Fuyang Normal College, Fuyang, China, in 2002, and the Ph.D. degree in information technology and engineering from Jiangnan University, Wuxi, China, in 2008.

He is currently a Professor with the School of Artificial Intelligence and Computer Science, Jiangnan University. He has visited the University of California-Davis and the Hong Kong Polytechnic University for more than two years. His current research interests include interpretable intelligence, uncertainty in artificial intelligence and their applications. He has authored or coauthored more than 100 research papers in international/ national journals.

Dr. Deng has served as an Associate Editor or Guest Editor of several international Journals, such as IEEE Trans. Emerging Topics in Computational Intelligence, *Neurocomputing*, and so on.

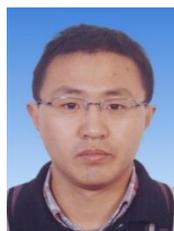

**Qingbing Sang** received the B.S. degree in computer science from the China University of Geosciences, Wuhan, China, in 1996, and the M.S. and Ph.D. degrees in pattern recognition from Jiangnan University, Wuxi, China, in 2005 and 2013, respectively. He was a Visiting Scholar with the LIVE Laboratory, the University of Texas at Austin, Austin, from August 2010 to August 2011.

He is currently an Associate Professor with the School of Artificial Intelligence and Computer Science, Jiangnan University. His research interests include computer vision and machine learning.

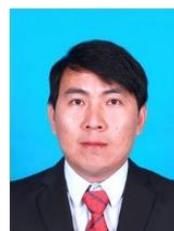

**Zhiyong Xiao** received the B.S. degree from Shandong University, China, in 2008, and the Ph.D. degree in Optics, Physics and Image Processing from the Ecole Central de Marseille, France, in 2014.

He is currently an Associate Professor with the School of Artificial Intelligence and Computer Science, Jiangnan University. He was an assistant research fellow at Fresnel Institute, CNRS France. His research interests include parallel computing, machine learning and artificial intelligence.

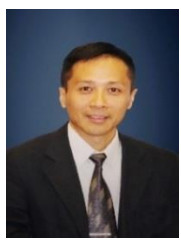

**Kup-Sze Choi** (M'97) received the Ph.D. degree in computer science and engineering from the Chinese University of Hong Kong, Hong Kong in 2004.

He is currently a Professor at the School of Nursing, Hong Kong Polytechnic University, Hong Kong, and the Director of the Centre for Smart Health. His research interests include virtual reality, artificial intelligence, and their applications in medicine and healthcare.

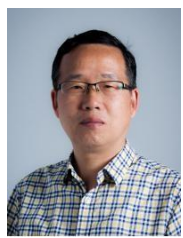

**Shitong Wang** received the M.S. degree in computer science from Nanjing University of Aeronautics and Astronautics, Nanjing, China, in 1987.

His research interests include artificial intelligence, neuro-fuzzy systems, pattern recognition, and image processing. He has published more than 100 papers in international/ national journals and has authored 7 books.